\documentclass{elsarticle}

\usepackage{url}
\usepackage{paralist}
\usepackage{multicol}
\usepackage{multirow}
\usepackage{algorithm}
\usepackage{algorithmic}
\usepackage{graphicx}
\usepackage{pstricks}
\usepackage{epsfig}
\usepackage{amsmath}
\usepackage{amssymb}
\usepackage{rotating}
\newtheorem{example}{Example}
\newtheorem{definition}{Definition}

\newcommand{\ie}{\textit{i.e.},~}
\newcommand{\eg}{\textit{e.g.},~}
\newcommand{\cf}{\textit{c.f.},~}
\newcommand{\wrt}{\textit{w.r.t.}~}

\makeatletter
\def\setliststart#1{\setcounter{\@listctr}{#1}%
  \addtocounter{\@listctr}{-1}}
\makeatother

\title{Formalising Hypothesis Virtues in Knowledge Graphs: A General 
Theoretical Framework and its Validation in Literature-Based Discovery 
Experiments\tnoteref{t1}}
\tnotetext[t1]{This work has been supported by the ``KI2NA'' 
project funded by Fujitsu Laboratories, Limited in collaboration with Insight, 
NUI Galway. We also greatly appreciate comments of Pierre-Yves Vandenbussche 
who helped us to refine the presentation of the article.} 

\author[vn]{V\'it Nov\'a\v{c}ek}
\address[vn]{
Insight @ NUI Galway (formerly known as DERI)\\
IDA Business Park, Lower Dangan, Galway, Ireland\\
\ead{vit.novacek@insight-centre.org}}

\begin{document}


\begin{abstract}
We introduce an approach to discovery informatics that uses so called knowledge
graphs as the essential representation structure. Knowledge graph is an 
umbrella term that subsumes various approaches to tractable representation of 
large volumes of loosely structured knowledge in a graph form. It has been 
used primarily in the Web and Linked Open Data contexts, but is applicable to 
any other area dealing with knowledge representation. In the perspective of 
our approach motivated by the challenges of discovery informatics, knowledge 
graphs correspond to hypotheses. We present a framework for formalising so 
called hypothesis virtues within knowledge graphs. The framework is based on a 
classic work in philosophy of science, and naturally progresses from mostly 
informative foundational notions to actionable specifications of measures 
corresponding to particular virtues. These measures can consequently be used 
to determine refined sub-sets of knowledge graphs that have large relative 
potential for making discoveries. We validate the proposed framework by 
experiments in literature-based discovery. The experiments have demonstrated 
the utility of our work and its superiority \wrt related approaches. 
\end{abstract}


\maketitle

\section{Introduction}\label{sec:introduction}

Ever since the dawn of computer age, researchers have been intrigued by the 
possibility of automating the process of discovery~\cite{newell1959processes}.
Today, the field of discovery informatics is getting more relevant than ever
before. The large amounts of data that are being made openly available for 
anyone to explore have an immense potential for making new discoveries, and
solutions that would enable this are highly sought 
after~\cite{honavar2014promise}. 

Knowledge graphs are one of the most universal ways of representing actionable, 
data-driven knowledge at large scale~\cite{dong2014knowledge}. They represent 
knowledge as relationships (edges) between items of interests (vertices), with 
the possibility of adding additional annotations representing for instance 
multiple relationship types (\ie predicates). Such a representation has many
advantages like universal applicability and a wealth of well-founded methods
for analysing graph structures. Yet the full potential of knowledge graphs for 
practical applications in knowledge discovery is still largely to be 
explored~\cite{dong2014knowledge}.

The motivation of the presented work is two-fold. Firstly, we want to propose 
a general framework for defining features of knowledge graphs that can 
determine which parts of the graphs have highest potential for making 
discoveries. We believe that this can facilitate the process of semi-automated
knowledge discovery in domains that have a lot of data available in graph-like 
format, but suffer from high redundancy and noise (\eg World Wide Web, social 
networks or biological pathway databases).

The second motivation is more practical. In our previous 
work~\cite{novacek2014peerj}, we addressed the problem of extracting simple
knowledge graphs from biomedical texts. The graphs were then used for so 
called machine-aided skim reading -- high-level navigation of a specific 
domain represented by a textual corpus which was assumed to facilitate the
discovery process. Indeed, even highly experienced domain experts were able to 
discover new and relevant facts using the prototype system. However, the 
results also contained some noise and connections that were correct, but rather
obvious and/or uninteresting. This motivated the validation experiments 
presented here, which demonstrate that our framework for formalising hypothesis
virtues can tackle the problems of noise, redundancy and obviousness in 
knowledge graphs automatically extracted from texts.

Our approach consists of formalising features applicable to ranking knowledge 
graphs (or their partitions) based on their potential for making discoveries. 
This can be used for instance for decomposing knowledge graphs into atomic 
subgraphs and consequent construction of a graph that has higher ``discovery 
potential'' than the original one. The formalisation is based on 
widely accepted hypothesis virtues studied in philosophy of 
science~\cite{quineullian1978webofbelief}. Examples of virtues are 
refutability or generality -- a good scientific hypothesis has to be 
falsifiable and should also provide explanations of phenomena outside of its
original scope. We present general conditions for each of the virtues and 
proceed with defining specific measures that conform to these conditions and 
can be efficiently implemented. 

The validation of the approach was performed in the context of literature-based
discovery~\cite{smalheiser2012literature}. We extracted knowledge graphs from 
two {\it de facto} standard biomedical corpora traditionally used in 
evaluation of literature-based discovery tools. For that we used a very simple
and domain-agnostic method that extracts statistically significant 
co-occurrence relationships. We opted for such a solution to demonstrate the
universal applicability of our approach. From these basic graphs, we 
constructed refined ones using a genetic algorithm that utilises the hypothesis 
virtue measures in the fitness function. The refined graphs were analysed 
according to the evaluation measures used in the literature-based discovery 
field and compared to related works. The results of the validation were 
positive, as we outperformed the state of the art in most respects. Moreover, 
we discovered relevant relationships that have not been covered by any related 
automated system or manual study. This demonstrates 
the practical utility of our approach.

Our main contributions are as follows. We have proposed a novel theoretical 
framework for extensible definition of measures that can be used to analyse 
the discovery potential of knowledge graphs. We have defined specific measures 
applicable especially to refinement of knowledge graphs automatically 
extracted from texts. We have implemented an evolutionary method for 
refinement of the automatically extracted knowledge graphs that is applicable 
out-of-the-box to any domain where English texts are available. We have 
demonstrated the practical relevance of the presented research by a successful 
experimental validation in the field of literature-based discovery. Last but 
not least, we have provided a data package containing a prototype 
implementation of our approach, results and other data necessary for the 
replication of our experiments.

The rest of the article is organised as follows. Section~\ref{sec:virtues} 
presents the general framework for formalising the hypothesis virtues in the
context of knowledge graphs. Section~\ref{sec:measures} then introduces 
actual measures that follow the general requirements of the hypothesis virtue 
formalisations. Our approach is experimentally validated in 
Section~\ref{sec:evaluation}. The section describes the evolutionary refinement
of knowledge graphs extracted from texts and elaborates on the experiments in
literature-based discovery. Related approaches are discussed in 
Section~\ref{sec:related}. Finally, we conclude the article and outline our 
future work in Section~\ref{sec:conclusion}.


\section{Formalising Hypothesis Virtues}\label{sec:virtues}

The foundations of the presented work are built 
on~\cite{quineullian1978webofbelief}, a classic work in philosophy of science. 
The work introduces five virtues of hypothesis: conservatism, modesty, 
simplicity, generality and refutability. These virtues present a comprehensive 
compilation of the philosophical treatments of discovery ranging from 
antiquity to modern analytical philosophy, and have been frequently used as a 
reference for determining quality of hypotheses in science.

According to~\cite{quineullian1978webofbelief}, the virtue of 
{\it conservatism} reflects the fact that good hypothesis usually
makes rather conservative claims. This is to minimise the risk of error by 
reaching too far from the state of the art in one step (even though the 
combination of the particular conservative claims may go very far after all, 
indeed). {\it Modesty} is related to conservatism -- a hypothesis A is more
modest than A and B (since A and B entails A), and a more modest hypothesis is 
considered better as it minimises the risk of wrong and/or redundant claims. 
The {\it simplicity} virtue posits that a good hypothesis should simplify our
view of the world by making new claims about it, even though the claims 
themselves may actually be quite complex. The {\it generality} virtue is
related to the predictive power of hypothesis -- the more phenomena (that have
perhaps not even been considered originally) it can explain, the better it is.
Finally, {\it refutability} means that a hypothesis should be falsifiable in
as obvious manner as possible. This is a factor of utmost importance, as 
discussed in arguably the most influential work on this 
topic~\cite{popper2005logic}.

In the following, we first define the notions of hypotheses and their claims in
the context of knowledge graphs (Section~\ref{sec:virtues.preliminaries}) and 
then continue with formalising the five virtues 
(Section~\ref{sec:virtues.formalising}). 

\subsection{Preliminaries}\label{sec:virtues.preliminaries}

First we define a universe -- a general knowledge graph within which 
particular hypotheses may be defined.
\begin{definition}\label{def:universe}
A universe graph $U$ is a tuple $(V_U, E_U, \Lambda_V,\Lambda_E)$ where $V_U$ 
is a set of vertices, $E_U \subseteq V_U \times V_U$ is a set of edges and
$\Lambda_V, \Lambda_E$ are sets of labeling maps (\ie morphisms) that 
associate values with the universe vertices and edges, respectively.
\end{definition}
The labeling maps can, for instance, assign predicate types to edges in
semantic networks, assert vertex types like class or individual in ontology 
knowledge graphs, or associate confidence weights with edges of automatically 
extracted knowledge graphs. Such a definition can accommodate a broad range of 
knowledge graphs with varying levels of semantic complexity, while keeping the 
basic structure still compatible with the analysis methods introduced here.
The universe can be either directed or undirected. The experiments presented 
in this article deal with an undirected universe and therefore we assume 
undirected graphs in the following unless explicitly stated otherwise.

A hypothesis in a universe is defined as follows.
\begin{definition}\label{def:hypothesis}
A hypothesis $H = (V_H,E_H,\Lambda_V^H,\Lambda_E^H)$ is a subgraph of the 
universe $U$ such that $V_H \subseteq V_U, E_H \subseteq E_U$ and
$\forall \lambda_V^H \in \Lambda_V^H \; \exists \lambda_V \in \Lambda_V.\;
\lambda_V^H \subseteq \lambda_V,\; \forall \lambda_E^H \in \Lambda_E^H \;
\exists \lambda_E \in \Lambda_E.\; \lambda_E^H \subseteq \lambda_E$.
\end{definition}
The second defining condition of the hypothesis subgraph means that any 
specific labeling map employed by a hypothesis has to be subsumed by a map
defined in the universe. This ensures that the universe is closed \wrt possible
interpretations of the hypotheses existing within it.

Most of the hypothesis virtues critically depend on what a claim of a 
hypothesis is, and therefore we need to define that as well.
\begin{definition}\label{def:claim}
A claim of a hypothesis $H$ is a simple (\ie acyclic) path in the graph $H$.
\end{definition}
Such a definition presents arguably the most universal view on what a 
particular knowledge graph may express. No matter what the actual semantics of 
the relationships in a hypothesis graph are, one can always study what they 
claim at least in terms of connections of vertices by means of edges, \ie 
paths (we will use the terms path and claim interchangeably in the rest of the 
article). This makes our approach applicable to any type of knowledge graph.

Note that one practical implication of the last definition is that we can 
consider only connected graphs as hypotheses -- if there is no path between 
two vertices, no claim is being made about them and they should thus be parts 
of different hypotheses. This is partly related to the open/closed world 
assumption dichotomy. The fact there is no connection between vertices does 
not mean no such connection can exist, it only means nothing is known about it 
in the context of the given knowledge graph.

The final preliminary definition concerns all claims possibly made by a 
hypothesis.
\begin{definition}
A claim set of a hypothesis $H$ is the set $\Pi(H)$ of all simple paths in 
the corresponding graph. A claim volume of $H$ is the size of its claim set,
\ie $|\Pi(H)|$.
\end{definition}
The claim volume can be very large and is hard to compute even for relatively 
small graphs~\cite{valiant1979complexity}. Also, it is not realistic to
expect every possible path in a knowledge graph to convey a meaningful claim. 
Therefore in practice, it is convenient to restrict the claim set to a more
manageable size based on case-specific heuristics. However, the maximal 
possible number of claims is apt as a theoretical notion for describing 
general knowledge graphs without further information about their domain
and more complex semantics.

\subsection{Formalising the Virtues}\label{sec:virtues.formalising}

The following five sections present formalisations of the particular hypothesis
virtues using the preliminary notions introduced above. Note that we provide 
general guidelines for measuring the virtues first, giving minimalistic set of 
conditions the measures should satisfy. Detailed examples of specific measures 
facilitating literature-based discovery are discussed in 
Sections~\ref{sec:measures} and~\ref{sec:evaluation}.

\subsubsection{Conservatism}\label{sec:virtues.formalising.conservatism}

Conservative claims should make small steps in a particular direction,
however, the combination of the steps can potentially be quite radical 
(\ie far-reaching). The conservatism of a path in a hypothesis $H$ can be 
measured by a function $f: \Pi(H) \rightarrow \mathbb{R}$ that satisfies the 
following conditions:
\begin{enumerate}
  \item Assuming a metric $\delta: V_U \times V_U \rightarrow \mathbb{R}$ on 
  the vertices in the universe graph, the function $f$ applied to a path 
  $p = (v_1,v_2,\dots,v_{|p|})$ is negatively correlated\footnote{Here and in 
  the following, we use broad notions of positive and negative correlation. 
  They are meant to generalise the respective notions of proportionality and 
  inverse proportionality to possibly non-linear, non-algebraic or statistical 
  relationships that may be specific to particular applications.} with the 
  $g(\{\delta(v_1,v_2),\delta(v_2,v_3),\dots,\delta(v_{|p|-1},v_{|p|})\})$ 
  value, where $g: 2^\mathbb{R} \rightarrow \mathbb{R}$ is an aggregation 
  function (\eg sum, mimimum, maximum or arithmetic mean).
  \item If radical claims are preferred, then there is an additional 
  requirement for $f$ being positively correlated with the $\delta(v_1,
  v_{|p|})$ value. 
\end{enumerate}


The conservatism of the whole hypothesis $H$ is computed by aggregating all
path conservatism measures across the $\Pi(H)$ set. The higher the 
aggregate value, the larger the conservatism. Due to the complexity of 
enumerating the $\Pi(H)$ set, practical conservatism measures can target only 
a subset of all possible paths. For instance, a set of shortest paths between 
all vertices in $H$ \wrt the $\delta$ edge labeling is a viable option as it 
is comparatively easier to compute and already satisfies condition 1. if sum
is used as an aggregation function.

\subsubsection{Modesty}\label{sec:virtues.formalising.modesty}

Let us refer by $H_{\omega}$ to the complete graph corresponding to a 
hypothesis $H$ (\ie a graph with an edge between any two vertices in $V_H$). 
Then the modesty of $H$ can be defined as 
$$
\frac{|\Pi(H_{\omega})|}{|\Pi(H)|}.
$$
This number reflects the ratio between all possible claims about the entities 
covered by $H$ and the actual number of claims being made. The higher the 
ratio, the larger the modesty (a modest hypothesis minimises the number of 
claims made in relation to the number of claims that can possibly be made).

As mentioned before, computing the number of all simple paths in a graph is 
extremely difficult in general. Therefore in practice, approximations of the
modesty measure are necessary. The approximations, however, should be
monotonic \wrt the ideal modesty measure: assuming $f,g$ as the ideal and
approximate modesty measures, respectively, then $g(H) > g(I)$ if and only if 
$f(H) > f(I)$ for any two hypotheses $H,I$. 

\subsubsection{Simplicity}\label{sec:virtues.formalising.simplicity}

For this virtue, we use the dual notion of complexity which has been 
extensively studied in the context of graphs~\cite{e14030559}. A good 
hypothesis should simplify our view of the world despite of possibly being 
locally complex~\cite{quineullian1978webofbelief}. In order to formalise this 
intuition, let us assume the simplicity of a graph is measured by a function 
$f: \mathcal{G_U} \rightarrow \mathbb{R}$, where $\mathcal{G_U}$ is a set of 
all graphs conceivable in the universe $U$. The function $f$ should satisfy 
these conditions:
\begin{enumerate}
  \item Given a hypothesis graph $H$ and a graph complexity measure 
  $c: \mathcal{G_U} \rightarrow \mathbb{R}$, $f$ is positively correlated with 
  the expression
  $$
  \frac{c(U \setminus H)}{c(U)}
  $$
  which reflects the universe simplification rate \wrt to the 
  hypothesis\footnote{From here on, we use the set-theoretic operators for 
  graphs as a convenience notation for the operations applied on the 
  corresponding vertex and edge sets in the actual tuple representations of 
  the graphs. The labeling sets of the result are assumed to be $\Lambda_V, 
  \Lambda_E$, \ie the universe ones, unless specified otherwise.}.
  \item If locally complex hypotheses are preferred, then the function $f$ 
  is also required to be positively correlated with the value $c(H)$.
\end{enumerate}
Strictly speaking, the rate in the first condition should also be higher than 
$1$ in order for the hypothesis to make the universe actually simpler, but 
practical applications may relax that requirement and just rank the hypotheses 
based on the measure. 

\subsubsection{Generality}\label{sec:virtues.formalising.generality}

Generality can be quantified as a number of explanations (\ie claims) the
hypothesis $H$ can provide for `out-of-scope' phenomena (\ie vertices) in the 
$U \setminus H$ graph. This can be expressed as
$$
g(\{f(u) | u \in V_U \setminus V_H\}),
$$
where the function $g: 2^\mathbb{R} \rightarrow \mathbb{R}$ is an aggregation 
(like sum or arithmetic mean) over all vertices that are out of the $H$ scope. 
The function $f: V_U \rightarrow \mathbb{R}$ is required to be positively 
correlated with the $h(\{|\{v|v \in p \wedge v \in V_H\}|\;|\;p \in 
\Pi_u(U)\})$ value, where $h: 2^\mathbb{R} \rightarrow \mathbb{R}$ is another
aggregation function and $\Pi_u(U)$ is a set of all simple paths in the 
universe $U$ that start in the vertex $u$. 

The generality definition reflects the basic intuition that the higher the 
number of $H$ vertices on paths explaining phenomena outside of $H$, the 
higher the generality of $H$. As the numbers of simple paths can be difficult 
to compute even if limited to paths starting in single nodes, approximations 
of this measure are needed for implementations again. Similarly to the modesty 
condition, we require the approximations to be monotonic \wrt the ideal 
generality measure. 

\subsubsection{Refutability}\label{sec:virtues.formalising.refutability}

Refutability can be seen as a quantification of:
\begin{inparaenum}[1)]
  \item the easiness with which the claim volume $|\Pi(H)|$ of a particular 
  hypothesis graph $H$ can be reduced;
  \item the rate of the reduction.
\end{inparaenum}
The atomic part of the process of refutation in the context of knowledge
graphs is an invalidation, \ie removal, of a vertex. Let us assume a 
decreasing ranking $R: \mathbb{N} \rightarrow V_U$ of the vertices in $H$ 
based on the number of simple paths that no longer exist in the graph after 
the vertex removal. Then we can define a top-k refutability as 
$$
\frac{|\Pi(H)|}{|\Pi(H)| + \sum_{i=1}^k |\Pi(H/R(i))|},
$$
where $H/R(i)$ is a graph resulting from removal of the first vertex in the 
ranking $R$ from the graph $H/R(i-1)$. We assert $H/R(0) = H$ by definition. 
The lower the number of paths still existing after removing the top vertex 
according to $R$, the higher the refutability. The $|\Pi(H)|$ expression is 
added to the denominator to avoid potential division by zero, and also to 
normalise the measure value.

Note that for growing $k$ values, the top-k refutability generally converges
to similar values for any given set of hypotheses as the measure is relative 
to the total number of paths in the graph. Therefore it is practical to use 
the measure with rather low $k$ values, perhaps even as low as $1$ which 
measures the rate of refutability in a single vertex removal step. 
Additionally, the ideal measure is difficult to compute and approximations are 
required in practice again. In particular, one can approximate the $\Pi$ 
function in the vertex ranking and refutability definition with one that is 
monotonic \wrt it.


\section{Specific Virtue Measures}\label{sec:measures}

In this part, we introduce specific instances of hypothesis virtue measures 
following the general formalisation presented before. First we give an
example of a universe and a couple of associated structures in 
Section~\ref{sec:measures.sample}. These will be used for running examples
illustrating the measure details in Section~\ref{sec:measures.definition}.
Finally, Section~\ref{sec:measures.combination} describes how to use the 
measures in concert. 

\subsection{Sample Universe}\label{sec:measures.sample}

\begin{figure}[ht]
\center
\scalebox{0.5}{\includegraphics{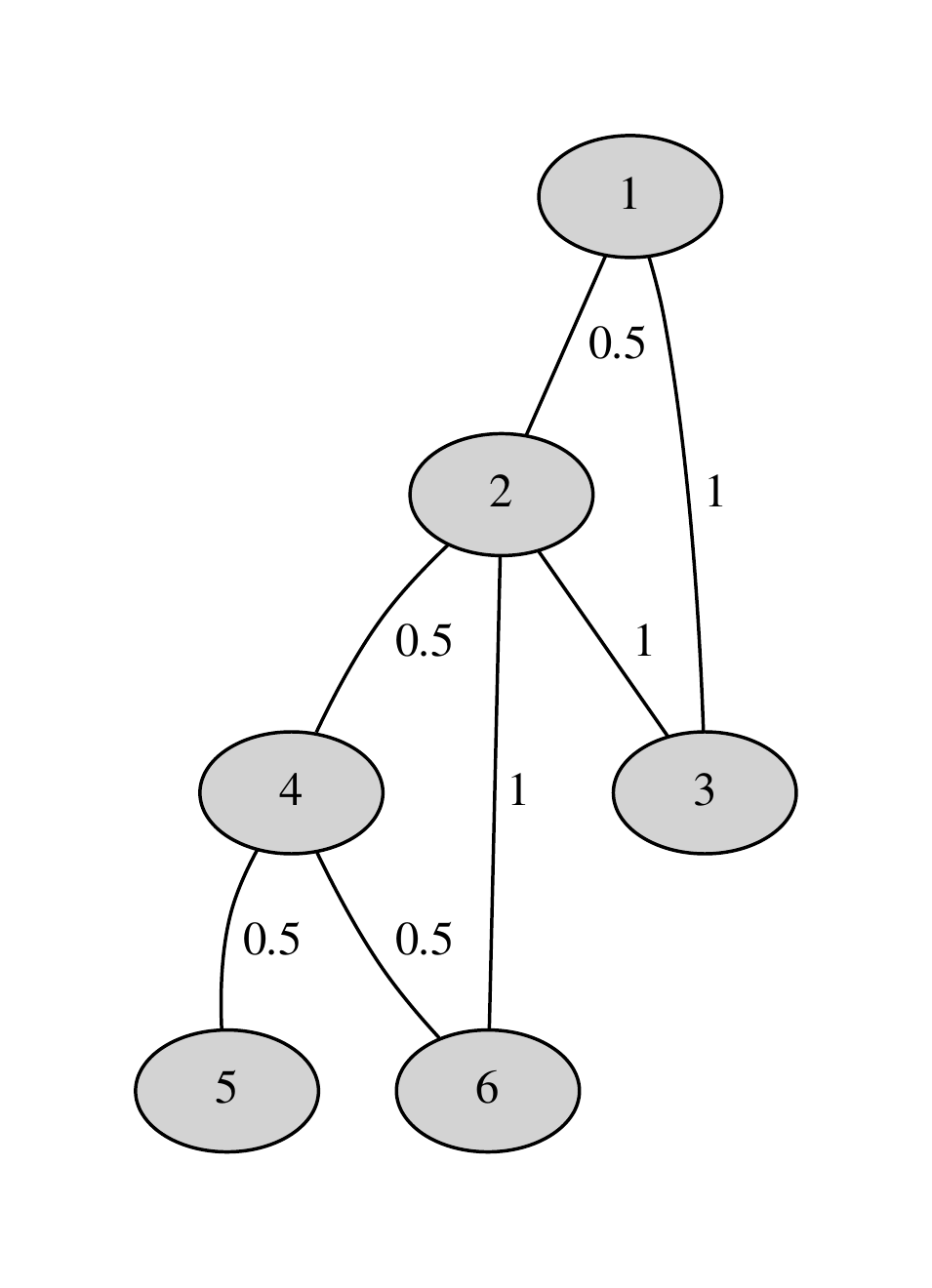}}
\caption{Sample universe graph $U$}\label{fig:toy_graph}
\end{figure}

The examples throughout this section are all based on an illustrative universe 
graph $U$ depicted in Figure~\ref{fig:toy_graph}.
The graph features real-valued edge labels in the $(0,1]$ interval that 
represent confidence weights of the edges (the higher the label the higher the 
expected degree of association between the corresponding vertices). These edge 
labels are used when constructing several auxiliary resources from the graph. 
There are no specific types of edges (\ie predicates) in the examples since in 
the experiments reported in this article, we focus only on one type of 
relationship based on automatically extracted co-occurrence statements.

First of all, we need to define a metric on the vertices. The most 
straightforward option without any background knowledge on the graph is to use 
its (weighted) adjacency matrix for constructing characteristic context vectors
for every vertex. The vectors can then be used for computing the actual metric.

The adjacency matrix $A_U$ of $U$ is presented in Table~\ref{tab:adj_matrix}.
\begin{table}[ht]
\footnotesize
\center
\begin{tabular}{l|cccccc}
   &  1  &  2  &  3  &  4  &  5  &  6  \\ \hline
1  &  0  & 0.5 &  1  &  0  &  0  &  0  \\
2  & 0.5 &  0  &  1  & 0.5 &  0  &  1  \\
3  &  1  &  1  &  0  &  0  &  0  &  0  \\
4  &  0  & 0.5 &  0  &  0  & 0.5 & 0.5 \\
5  &  0  &  0  &  0  & 0.5 &  0  &  0  \\
6  &  0  &  1  &  0  & 0.5 &  0  &  0  \\
\end{tabular}
\caption{Weighted adjacency matrix $A_U$}\label{tab:adj_matrix}
\end{table}
The context vector $\mathbf{x}$ for a vertex $x$ is the row (or column, as the
graph is undirected) corresponding to $x$ in the adjacency matrix $A_U$. 
Using the context vectors, we can define the Euclidean distance (\ie a metric)
on the vertices as $\delta(x,y) = \sqrt{\sum_{i = 1}^n(x_i-y_i)^2}$ where 
$x_i,y_i$ correspond to the $i$-th elements of the $\mathbf{x},\mathbf{y}$
context vectors, respectively. The specific distances (up to 4-th decimal
point) between the universe vertices are given in Table~\ref{tab:dist_matrix}.
\begin{table}[ht]
\footnotesize
\center
\begin{tabular}{l|cccccc}
   & 1      & 2      & 3      & 4      & 5      & 6      \\ \hline
1  & 0      & 1.3229 & 1.5    & 1.2247 & 1.2247 & 1.2247 \\
2  & 1.3229 & 0      & 1.8708 & 1.5    & 1.5    & 1.8028 \\
3  & 1.5    & 1.8708 & 0      & 1.3229 & 1.5    & 1.118  \\
4  & 1.2247 & 1.5    & 1.3229 & 0      & 1      & 1      \\
5  & 1.2247 & 1.5    & 1.5    & 1      & 0      & 1      \\
6  & 1.2247 & 1.8028 & 1.118  & 1      & 1      & 0      \\
\end{tabular}
\caption{Distance matrix $D_U$}\label{tab:dist_matrix}
\end{table}

\begin{figure}[ht]
\center
\scalebox{0.5}{\includegraphics{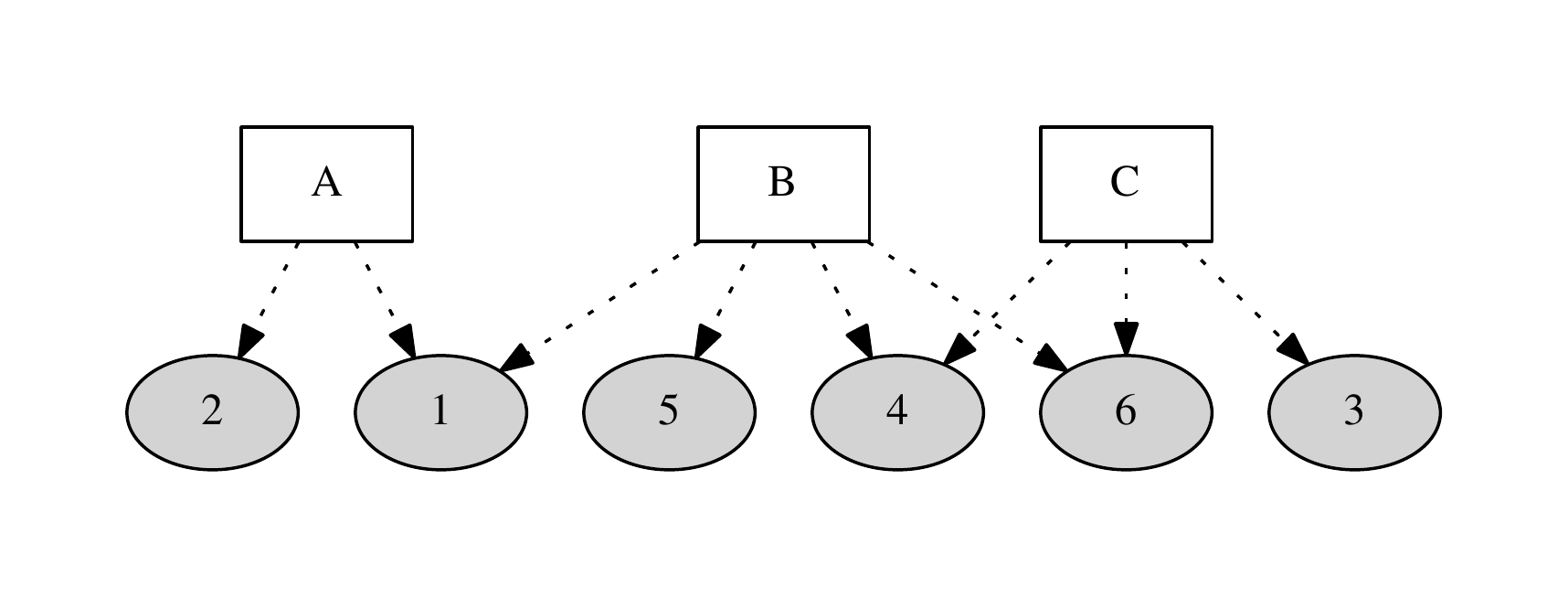}}
\caption{
Cluster structure of $U$}\label{fig:cluster_graph}
\end{figure}

The last auxiliary structure we will need in the following sections (namely
for defining complexity measures) is clustering of the vertices in $U$. An 
example of a possible clustering is given in Figure~\ref{fig:cluster_graph}.
It is an 
overlapping 
clustering that 
groups vertices with mutual distances below $1.5$ (in the actual implementation
of our approach, we use more sophisticated clustering method as explained in 
detail in Section~\ref{sec:evaluation.evolution.graph_construction}). 
The clustering contains three 
clusters $A: \{1,2\}, B: 
\{1,4,5,6\}, C: \{3,4,6\}$. 
Note that the clustering can either be computed from the universe graph itself
or provided externally (\eg in the form of an ontology that defines a taxonomy
upon the graph vertices).

\subsection{Measure Definitions}\label{sec:measures.definition}

Having introduced the sample universe, we can continue with the specific 
measure definitions which we use later on in the literature-based discovery 
experiments.

\subsubsection{Conservatism}\label{sec:measures.definition.conservatism}

Following the conditions provided in 
Section~\ref{sec:virtues.formalising.conservatism}, we define a specific 
instance of the hypothesis graph conservatism measure
$$
C(H) = \frac{1}{|\pi_s(H,\delta)|}\sum_{p \in \pi_s(H,\delta)}
\frac{\delta(v_1,v_{|p|})}{\sum_{i=1}^{|p|-1} \delta(v_i,v_{i+1})},
$$
where $\pi_s(H,\delta)$ is a set of all shortest paths in $H$ \wrt the 
Euclidean distance $\delta$ and $p = (v_1,v_2,\dots,v_{|p|})$ is a specific 
shortest path of length $|p|$. In other words, the $C$ measure is an 
arithmetic mean of the shortest path conservatism values where the path 
conservatism is computed as a fraction of the distance between the extreme 
vertices of the path and the path length\footnote{Note that if there is only 
one shortest path guaranteed to exist between any pair of vertices in the $H$ 
graph, then $|\pi_s(H,\delta)| = {|V_H| \choose 2}$ as $H$ is expected to be 
connected.}. 

The measure satisfies the condition 1. from 
Section~\ref{sec:virtues.formalising.conservatism} as it already focuses only
on paths with minimal aggregate distance between the consecutive vertices 
(assuming the sum aggregation). The condition 2. is satisfied as well. For any 
path $p$, $\delta(v_1,v_{|p|}) \leq \sum_{i=1}^{|p|-1} \delta(v_i,v_{i+1})$. 
The equality is achieved if and only if the context vectors of the consecutive 
vertices represent points that lie in a straight line, \ie maximise the 
distance between the extreme vertices of the path. Therefore the maximum value 
$1$ of the path conservatism measures is achieved exactly when the extreme 
distance is maximal.

\begin{example}\label{ex:conservatism}
In Figure~\ref{fig:hypotheses_graphs} there are three hypothesis graphs 
$E,F,G$ that exist in the universe $U$ described in 
Section~\ref{sec:measures.sample}. 
\begin{figure}[ht]
\center
\scalebox{0.5}{
  \includegraphics{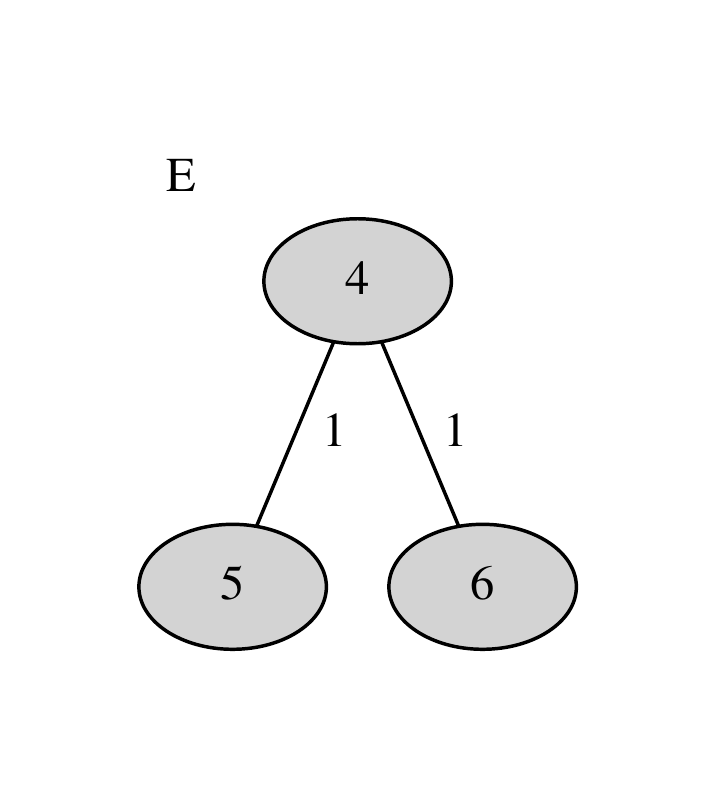}
  \includegraphics{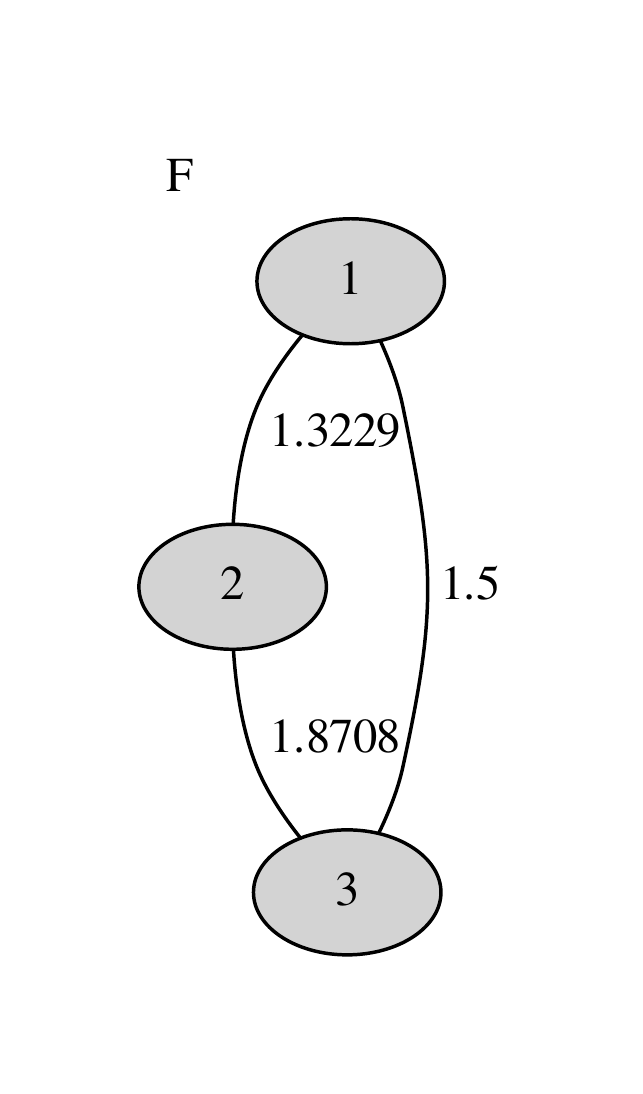}
  \includegraphics{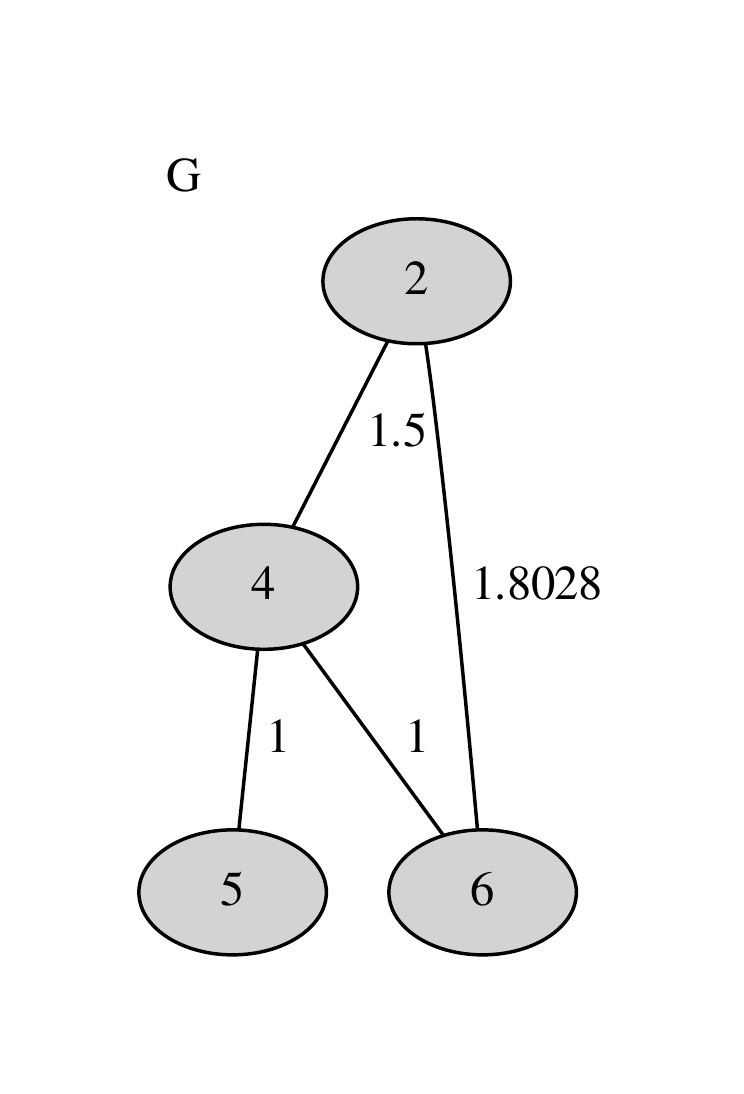}
}
\caption{Sample hypothesis graphs $E,F,G$}\label{fig:hypotheses_graphs}
\end{figure}
The edges are annotated with the Euclidean distance $\delta$ based on the 
vertex context vectors (see the examples in Section~\ref{sec:measures.sample} 
for details). 

The numbers of all shortest paths for the hypothesis graphs $E,F,G$ \wrt the 
distance $\delta$ are $3,3,6$, respectively. The conservatism measures of the
hypotheses are
$$
C(E) = \frac{1}{3}(\frac{1}{1} + \frac{1}{2} + \frac{1}{1}) = 0.8\overline{3},
$$
$$
C(F) = \frac{1}{3}(\frac{1.3229}{1.3229} + \frac{1.8708}{1.8708} + 
       \frac{1.5}{1.5}) = 1,
$$
$$
C(G) = \frac{1}{6}(\frac{1.5}{1.5} + \frac{1.8028}{1.8028} + 
                   \frac{1}{1} + \frac{1}{1} + \frac{1.5}{2.5} + 
                   \frac{1}{2}) = 0.85,
$$
therefore the hypotheses can be ranked in the 
$$
F \succ_C G \succ_C E
$$ 
order from the most to the least conservative 
one\footnote{From here on, we use convenience ordering relations $\succ_{X}$ 
for ranking the hypotheses in a decreasing order according to a specific 
measure $X$. $E \succ_X F$ if and only if $X(E) > X(F)$.}.
\end{example}

\subsubsection{Modesty}\label{sec:measures.definition.modesty}

As an approximation of the the ideal modesty measure presented in
Section~\ref{sec:virtues.formalising.modesty}, we use inverse density of the 
hypothesis graph
$$
M(H) = \frac{|V_H|(|V_H|-1)}{2|E_H|}.
$$
This function is much easier to compute than the ideal one and is monotonic 
\wrt it. Since the enumerators of both functions are fixed, we only need to 
show that the number of edges is monotonic \wrt number of all simple paths in
a hypothesis graph. This is quite easy -- increase in $|E_H|$ (\ie adding an 
edge) will cause $|\Pi(H)|$ to grow as well since adding an edge will result
in at least one new simple path in $H$, the edge itself. Conversely, if the set
$\Pi(H)$ grows, it means that edges had to be added to the $H$ graph as it is
the only way how the overall number of paths can be increased. 

\begin{example}\label{ex:modesty}
The number of edges in the $E,F,G$ graphs from Example~\ref{ex:conservatism} is
$2,3,4$, respectively, while the maximum possible number of edges in the 
corresponding complete graphs is $3,3,6$. Therefore the modesty values are
$$
M(E) = \frac{3}{2} = 1.5,\;M(F) = \frac{3}{3} = 1,\;M(G) = \frac{6}{4} = 1.5
$$
and the modesty ranking of the hypotheses is
$$
E \succ_M F,\;G \succ_M F.
$$
\end{example}

\subsubsection{Simplicity}\label{sec:measures.definition.simplicity}

As stated in Section~\ref{sec:virtues.formalising.simplicity}, we use the 
dual notion of complexity for measuring hypothesis simplicity. For the specific
instance of the measure, we employ Shannon's entropy that has been frequently 
used for graph complexity~\cite{e14030559}. To define the entropy, we utilise 
the clustering of the hypothesis graph vertices based on their context vectors.
Let us assume a vertex labeling $\gamma: V_U \rightarrow 2^L$ where $L$ is a set
of cluster identifiers. Then we can define a cluster association probability
$p(l,H)$ for a specific cluster $l \in L$ within a hypothesis $H$ as
$$
p(l,H) = \frac{|\{v|v \in V_H \wedge l \in \gamma(v)\}|}{|V_H|}.
$$
It is a probability that a randomly selected vertex from $H$ belongs to a 
cluster $l$. If we conceive clusters as higher-level topics the hypothesis
graph deals with, then the probability reflects the distribution of
the topics across the graph. The $p(l,H)$ values can be used for computing 
the cluster association entropy for a hypothesis $H$ as
$$
E(H) = -\sum_{l \in L} p(l,H) \log_2 p(l,H).
$$
It reflects the information value of the hypothesis' cluster structure --
the more ``unpredictably'' distributed clusters, the higher the complexity and
also the information value. This conforms to an intuitive assumption that 
hypotheses dealing with more topics representatively are more informative, \ie 
complex.

We define two simplicity measures that employ the cluster association entropy 
and satisfy the respective conditions introduced in 
Section~\ref{sec:virtues.formalising.simplicity}
$$
S_1(H) = E(H),\; S_2(H) = \frac{E(U \setminus H)}{E(U)}.
$$
We use both measures in the following to capture different aspects of 
simplicity simultaneously.

\begin{example}\label{ex:simplicity}
In Figure~\ref{fig:hypotheses_graphs_c} there are the three hypotheses graphs 
$E,F,G$ and the universe graph $U$ depicted again, but this time with cluster
annotations provided as vertex labels.
\begin{figure}[ht]
\center
\scalebox{0.5}{
  \includegraphics{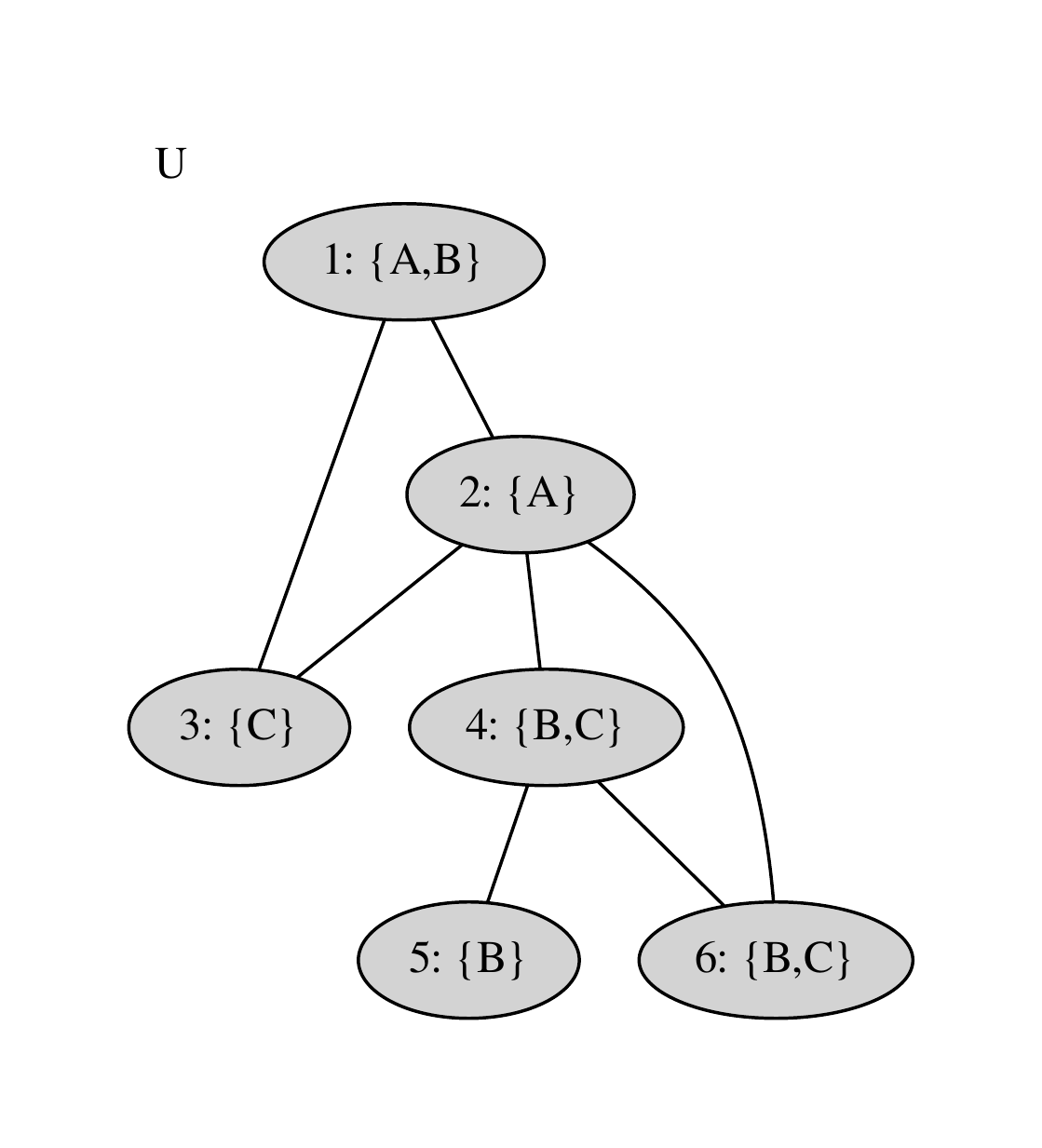}
}\\
\scalebox{0.5}{
  \includegraphics{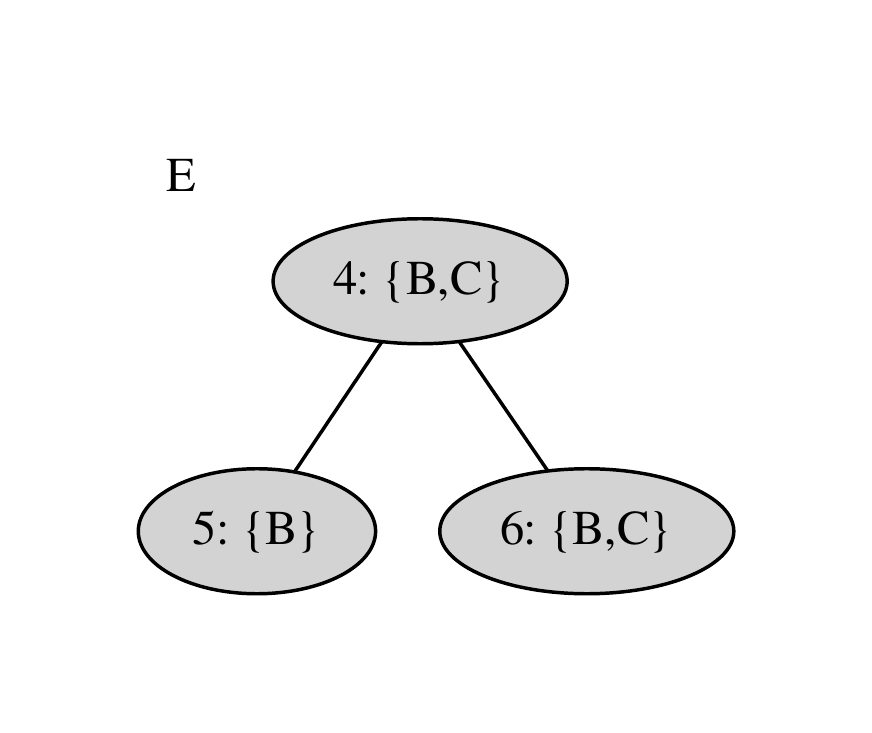}
  \includegraphics{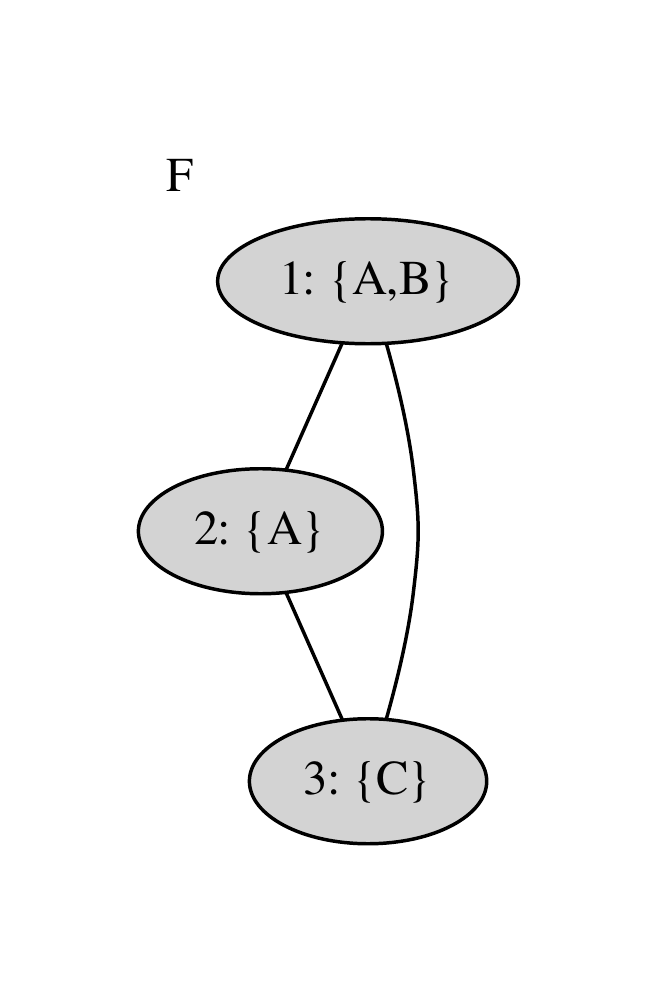}
  \includegraphics{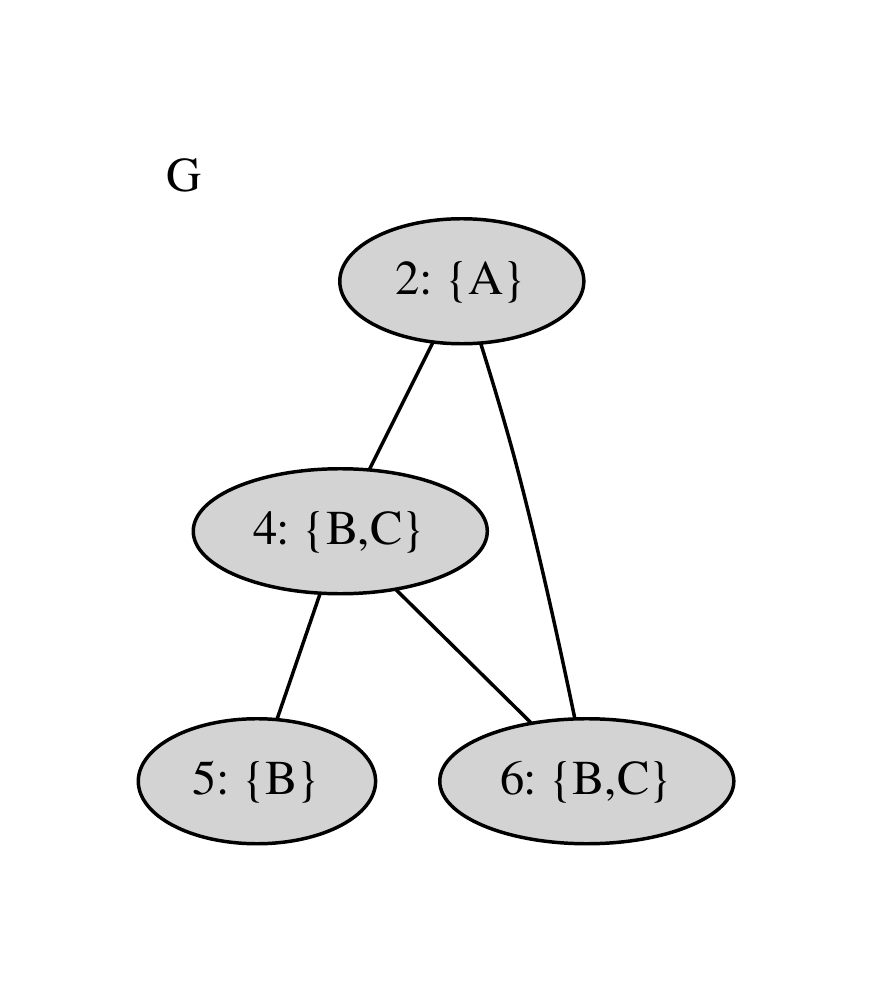}
}
\caption{Sample hypothesis graphs $E,F,G$}\label{fig:hypotheses_graphs_c}
\end{figure}
The cluster association probabilities for each graph are
$$
p(A,U) = \frac{1}{3},\; p(B,U) = \frac{2}{3},\; p(C,U) = \frac{1}{2},
$$
$$
p(A,E) = 0,\; p(B,E) = 1,\; p(C,E) = \frac{2}{3},
$$
$$
p(A,F) = \frac{2}{3},\; p(B,F) = \frac{1}{3},\; p(C,F) = \frac{1}{3},
$$
$$
p(A,G) = \frac{1}{4},\; p(B,G) = \frac{3}{4},\; p(C,G) = \frac{1}{4}.
$$
The entropies corresponding to these probabilities are
$$
E(U) \doteq 1.4183,\; E(E) \doteq 0.39,\; E(F) \doteq 2.78,\; E(G) \doteq 
1.3113,
$$
$$
E(U \setminus E) \doteq 2.78,
E(U \setminus F) \doteq 0.39,
E(U \setminus G) = 1.5.
$$
The hypothesis $F$ is the lowest-ranking no matter which function we use -- it 
has the lowest entropy and $E(U \setminus F) < E(U)$, therefore it makes the 
universe more complex. On the other hand, both $E,G$ increase the simplicity 
of the universe. If only local complexity of the acceptable hypotheses is
relevant (measure $S1$), then the final ranking is
$$
G \succ_{S1} E \succ_{S1} F
$$
since $E(G) > E(E)$. However, if the rate of simplifying the universe is more
important (measure $S2$), the ranking is
$$
E \succ_{S2} G \succ_{S2} F
$$
as 
$$
\frac{E(U \setminus E)}{E(U)} \doteq 1.9601 > 1.0576 \doteq 
\frac{E(U \setminus G)}{E(U)}.
$$
\end{example}

\subsubsection{Generality}\label{sec:measures.definition.generality}

To limit the potentially intractable number of paths in the ideal generality
formula introduced in Section~\ref{sec:virtues.formalising.generality}, we
apply two approximations in its specification. Firstly, we focus only on 
explanations for the universe vertices $V_A^H$ that are immediately adjacent 
to the measured hypothesis $H$. The set of edges that connect these vertices 
to $H$ can then be defined as $E_A^H = \{(u,v) | (u,v) \in E_U \wedge (u \in 
V_A^H, v \in V_H \vee u \in V_H, v \in V_A^H)\}$. The second approximation 
consists of focusing only on shortest paths \wrt the $\delta$ distance. The 
specific generality measure is then defined as
$$
G(H) = |\{p | p \in \pi_s(U,\delta) \wedge p_1 \in V_A^H \wedge 
p_2,p_3,\dots,p_{|p|} \in V_H\}|.
$$
The measure corresponds to the number of shortest paths that start in an 
adjacent vertex and connect it with vertices in the hypothesis graph $H$,
thus providing an explanation for it using only $H$. As the graphs 
$H$ are assumed to be connected, the measure can further be simplified as 
$G(H) = |E_A^H|(1+(V_H-1)) = |E_A^H||V_H|$ for graphs where only one shortest 
path exists between any two vertices.

The $G(H)$ measure uses sum aggregation as the $g$ function present in the 
general definition. The $f$ function that leads to the presented definition of 
$G(H)$ returns zero for any vertex from the $V_U \setminus V_H$ set that is 
not immediately adjacent to $H$. For other vertices, it returns the number of 
paths that provide explanation for them in $H$. This number is positively 
correlated with the number of vertices in $V_H$ as required in the general 
definition, since the number of paths leading from a vertex to other vertices 
in a connected graph $H$ is $|V_H|-1$ (or more if multiple shortest paths 
exist between some vertices). 

The restriction to the immediately adjacent vertices leads to a narrowly 
focused generality and helps to reduce combinatorial explosion resulting from 
taking the whole universe graph into account. The shortest path approximation 
is a reasonable limitation as these paths are more likely to be conservative 
explanations. It is not strictly monotonic \wrt the ideal generality measure,
though. If the number of shortest paths increases, then the number of all 
paths naturally has to be higher as well. The other direction is less obvious, 
and conditional. Assuming the number of all paths in a graph has increased, we 
have to show that there also has to be more shortest paths. This is not true 
in general -- if edges between distant vertices are added, they may not 
contribute to increasing the number of shortest paths. However, since the 
measure intuitively captures the notion of generality in the context of 
knowledge graphs and is easy to compute, we decided to relax the absolute 
monotonicity requirement for the sake of practicality.

\begin{example}\label{ex:generality}
The sets vertices adjacent to the $E,F,G$ hypotheses are
$$
V_A^E = \{2\},\; V_A^F = \{4,6\}, V_A^G = \{1,3\}
$$
and the corresponding sets of connecting edges are
$$
E_A^E = E_A^F = \{(4,2), (6,2)\}, \; E_A^G = \{(2,1), (2,3)\}.
$$
Since there is only one shortest path between any pair of vertices in our
example, the generality measures are
$$
G(E) = |E_A^E| \cdot |V_E| = 2 \cdot 3 = 6,\; 
G(F) = |E_A^F| \cdot |V_F| = 2 \cdot 3 = 6,\;
$$
$$
G(G) = |E_A^G| \cdot |V_G| = 2 \cdot 4 = 8
$$
and the resulting ranking is
$$
G \succ_G E,\; G \succ_G F.
$$
\end{example}

\subsubsection{Refutability}\label{sec:measures.definition.refutability}

Using the shortest paths approximation again, we define a specific refutability
measure as
$$
R_k(H) = \frac{|\pi_s(H,\delta)|}{|\pi_s(H,\delta)| + 
\sum_{i=1}^k |\pi_s(H/R(i),\delta)|}.
$$
Similarly to Section~\ref{sec:measures.definition.generality}, we consider 
only the 
shortest paths instead of all simple ones, which makes the computation of the
measure comparatively easier. Such an approximation is unfortunately not 
strictly monotonic as shown before, however, we believe that the practicality 
and intuitiveness of the measure outweighs the partial monotonicity violation.

For the ranking $R$ of the vertices in the $R_k(H)$ measure computation, we use
the betweenness centrality which is defined as
$$
c_B(v,G) = \frac{|\{p|p \in \pi_s(G,\delta) \wedge v \in 
p\}|}{|\pi_s(G,\delta)|},
$$
where $v$ is a vertex and $G$ is a graph. In other words, betweenness 
centrality of a vertex is the number of shortest paths passing though it 
divided by total number of shortest paths. The ranking $R$ ranks the vertices 
in a decreasing order based on their betweenness centrality. Such ranking 
generally does not mean that removal of a high-ranking vertex results in a
higher number of shortest paths disappearing when compared to a removal of a
lower-ranking vertex -- if the graph remains connected in both cases, the
number of shortest paths in it will be the same after removal of either node.
However, removing a vertex with higher betweenness centrality will result in
relative increase of the remaining paths' lengths. This can lead to a decrease
of the graph conservatism and thus also to a decrease of its overall value
\wrt the hypothesis virtues. Consequently, making a hypothesis weaker more 
quickly can be seen as refuting it more efficiently. We believe that this 
justifies the chosen ranking even though it means yet another relaxation of 
the general requirements\footnote{An alternative option that fully conforms to 
the requirements would employ simple paths instead of shortest ones and vertex
degree instead of betweenness centrality, however, such a solution can easily
become intractable.}.

\begin{example}\label{ex:refutability}
The sets of shortest paths \wrt the $\delta$ distance for the particular
hypothesis graphs are
$$
\pi_s(E,\delta) = \{(5,4), (5,4,6), (4,6)\},
$$
$$
\pi_s(F,\delta) = \{(2,1), (2,3), (1,3)\},
$$
$$
\pi_s(G,\delta) = \{(2,4), (2,4,5), (2,6), (4,5), (4,6), (5,4,6)\}.
$$
The corresponding vertex betweenness centralities are then
$$
c_B(4,E) = 1,\; c_B(5,E) = c_B(6,E) = 0.\overline{6},
$$
$$
c_B(1,F) = c_B(2,F) = c_B(3,F) = 0.\overline{6},
$$
$$
c_B(2,G) = c_B(5,G) = c_B(6,G) = 0.5,\; c_B(4,G) = 0.8\overline{3}.
$$
The top-1 refutability measure for the hypothesis $E$ can be computed as 
follows. The centrality-based ranking of the vertices places $4$ on the top,
therefore we remove it. The result is a disconnected graph consisting of
isolated vertices $5,6$ where no path exists anymore. The top-1 refutability 
measure of $E$ is thus
$$
R(E,1) = \frac{3}{3+0} = 1.
$$
Similarly, the top-1 refutability measures for the remaining two hypotheses 
(with arbitrary removal vertex selection for $F$ due to uniform centrality 
ranking) are
$$
R(F,1) = \frac{3}{3+1} = 0.75,\; R(G,1) = \frac{6}{6+1} = 0.\overline{857142}.
$$
The resulting refutability ranking of $E,F,G$ is
$$
E \succ_R G \succ_R F.
$$
\end{example}

\subsection{Combining the Measures}\label{sec:measures.combination}

The specific measures defined in the previous section can be used to rank the 
hypothesis graphs independently of each other as shown in the examples. 
However, practical applications will very often imply the necessity to compare 
hypotheses along all the measures. Lacking any {\it a priori} information on 
which measures may be more relevant for a particular application, we propose 
the following way of ordering the hypothesis graphs. 

Let $\mathcal{H} = \{H_1,H_2,\dots,H_n\}$ be the set of hypothesis graphs we 
wish to compare according to a set of measures $\mathcal{X} = \{X_1,X_2,\dots,
X_m\}$ of equal importance. Then we can construct an edge-labeled directed 
ranking multigraph $\mathcal{R} = (\mathcal{H},\mathcal{E} \subseteq 
\mathcal{H} \times \mathcal{H},\lambda: \mathcal{E} \rightarrow \mathcal{X})$. 
The multigraph's vertices are the hypotheses in $\mathcal{H}$. The edge set 
and the labeling function is constructed from the specific measure rankings so 
that $(H_i,H_j) \in \mathcal{E}, \lambda(H_i,H_j) = X_k$ if and only if 
there is a measure $X_k$ such that $H_i \succ_{X_k} H_j$. Using the ranking 
multigraph $\mathcal{R}$, we can define a combined ranking relation $\succ$ on
the set $\mathcal{H} \times \mathcal{H}$ as
$$
H_i \succ H_j \mathrm{\;if\;and\;only\;if\;} 
\frac{d_o(H_i,\mathcal{R})}{d_o(H_i,\mathcal{R}) + d_i(H_i,\mathcal{R})} > 
\frac{d_o(H_j,\mathcal{R})}{d_o(H_j,\mathcal{R}) + d_i(H_j,\mathcal{R})},
$$
where $d_i(H_x,\mathcal{R}), d_o(H_x,\mathcal{R})$ is the in-degree and 
out-degree of the vertex $H_x$ in the multigraph $\mathcal{R}$, respectively.
In plain words, the combined ranking relation $\succ$ orders the hypotheses
based on the relative magnitude of their superiority (out-degree) 
\wrt the specific ranking relations given by the 
measures. 

\begin{example}\label{ex:combination}
Figure~\ref{fig:rankings} shows the ranking multigraph corresponding to 
Examples~\ref{ex:conservatism}-\ref{ex:refutability}.
\begin{figure}[ht]
\center
\scalebox{0.5}{\includegraphics{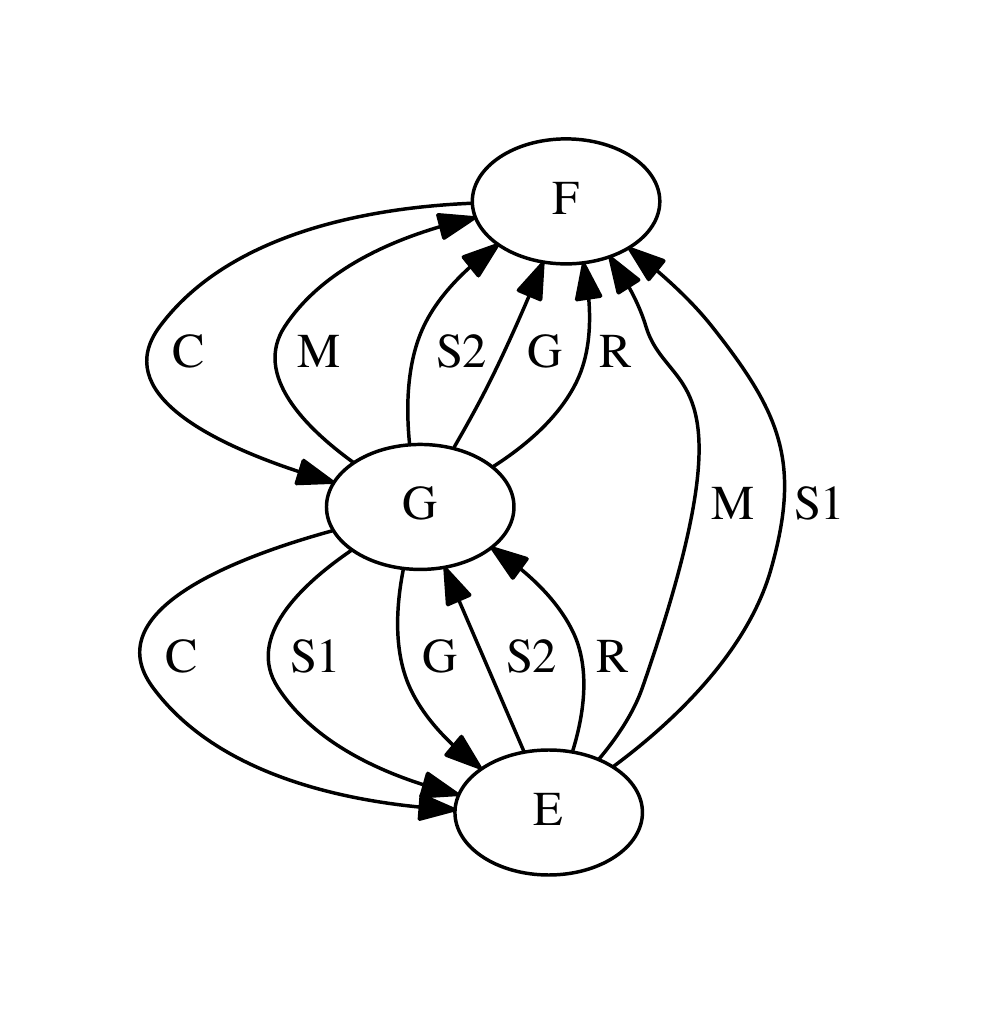}}
\caption{Ranking multigraph for $E,F,G$}\label{fig:rankings}
\end{figure}
A directed edge from vertex $X$ to $Y$ with a label $Z$ means that 
$X \succ_Z Y$. The in-degrees and out-degrees of $E,F,G$ in the ranking
graph are
$$
d_i(E) = 3,\; d_o(E) = 4,\;
d_i(F) = 6,\; d_o(F) = 1,\;
d_i(G) = 3,\; d_o(G) = 7,
$$
therefore
$$
G \succ E \succ F
$$
since
$$
\frac{7}{10} > \frac{4}{7} > \frac{1}{7}.
$$
\end{example}


\section{Experimental Validation}\label{sec:evaluation}

In order to validate the proposed formalisation of hypothesis virtues in the 
context of knowledge graphs, 
we chose to follow-up on our work presented in~\cite{novacek2014peerj} where
we addressed automated extraction of conceptual networks from biomedical 
literature. The work deals with extraction of co-occurrence and similarity 
relationships from abstracts available on PubMed (\cf 
\url{http://www.ncbi.nlm.nih.gov/pubmed}) and consequent indexing, querying 
and navigation of the networks in a knowledge discovery scenario.

As we have shown in~\cite{novacek2014peerj}, the automatically extracted
networks can already provide useful insights even for experts in the field, 
however, they still contain some noise and irrelevant and/or obvious 
information. Tackling this challenge has been the main practical motivation for
the research presented in this article. We believe we can use our approach to 
identify portions of the automatically extracted graphs that can not only 
provide general overview of the domain with less noise, but also isolate 
valid relationships that are surprising for experts. This can ultimately lead 
to more efficient machine-aided discovery applications.

In our validation experiments, we utilise the scenarios, data sets and 
evaluation methodologies elaborated within the field of literature-based 
discovery which we introduce in Section~\ref{sec:evaluation.lbd} below. 
Section~\ref{sec:evaluation.evolution} is the methodological core of this part.
It presents an evolutionary approach to the refinement of automatically 
extracted knowledge graphs using the hypothesis virtue measures. 
Section~\ref{sec:evaluation.experiments} describes the data sets and methods
we use for the experimental evaluation. Finally, 
Section~\ref{sec:evaluation.results} discusses the results of the experiments. 

Note that we have implemented our approach and the experiments reported in 
this section using a Python prototype available under the GPL free software 
license. The corresponding code, experimental data and results are available at 
\url{http://skimmr.org/hyperkraph/}\footnote{HYPERKRAPH is a general name we 
use for the ongoing implementation of prototypes based on the presented 
research. It stands for {\it HYPothEsis viRtues in Knowledge gRAPHs}.}. 
Detailed README documentation on the implementation and data is provided as a 
part of the respective archives hosted at the referenced URL.

\subsection{Literature-Based Discovery}\label{sec:evaluation.lbd}

The field of literature-based discovery is widely considered to stem from the 
work~\cite{swanson1986fish}. Based on~\cite{swanson1986fish} and a follow-up
article~\cite{swanson1987migraine}, the work~\cite{stegmann2003hypothesis} 
introduced the notion of Swanson linking -- connecting two pieces of knowledge 
in isolated documents A and B using concepts from intermediate documents (C)
that are directly or indirectly related to A and B. Surveys of recent works 
addressing this problem 
are provided 
in~\cite{deBruijn20027,preiss2012towards,smalheiser2012literature}.

The application of our framework to refining knowledge graphs automatically 
extracted from literature is closely related to literature-based discovery. 
Our goal is to generate a set of graphs that reflect relationships between 
terms in literature and are optimised \wrt hypothesis virtues. Such a 
structure can very straightforwardly facilitate the process of finding 
``interesting'' links between isolated concepts via intermediates, which is 
the key problem of literature discovery. Therefore we can use the standard 
approaches and man-made ``gold standard'' discoveries from that field to 
experimentally validate our approach in an established application scenario. 

\subsection{Evolutionary Refinement of Automatically Extracted Knowledge 
Graphs}\label{sec:evaluation.evolution}

The basic assumption we use for validating our framework is that applying the 
hypothesis virtue measures to refining graphs extracted from literature will 
facilitate literature-based discovery tasks better than the unrefined graphs. 
To verify this, we have to tackle the graph refinement first. The key question
is:
{\it Given a knowledge graph based on statements automatically extracted from 
text, how can we refine it so that only the parts of the graph that have 
comparatively high hypothesis virtue measures remain?}

This is essentially an optimisation problem in which we know how to tell 
whether a solution X is better than Y, but we do not know much about what the
actual solutions are and how the main knowledge graph is (or should be) 
composed of them. Such problems can quite efficiently be tackled by 
evolutionary computing~\cite{evolcomp}. In the rest of this section, we
describe a specific algorithm for evolutionary refinement of knowledge graphs.

\subsubsection{Extracting a Universe Graph from 
Texts}\label{sec:evaluation.evolution.graph_construction}

\begin{figure}
\center
\scalebox{0.5}{\includegraphics{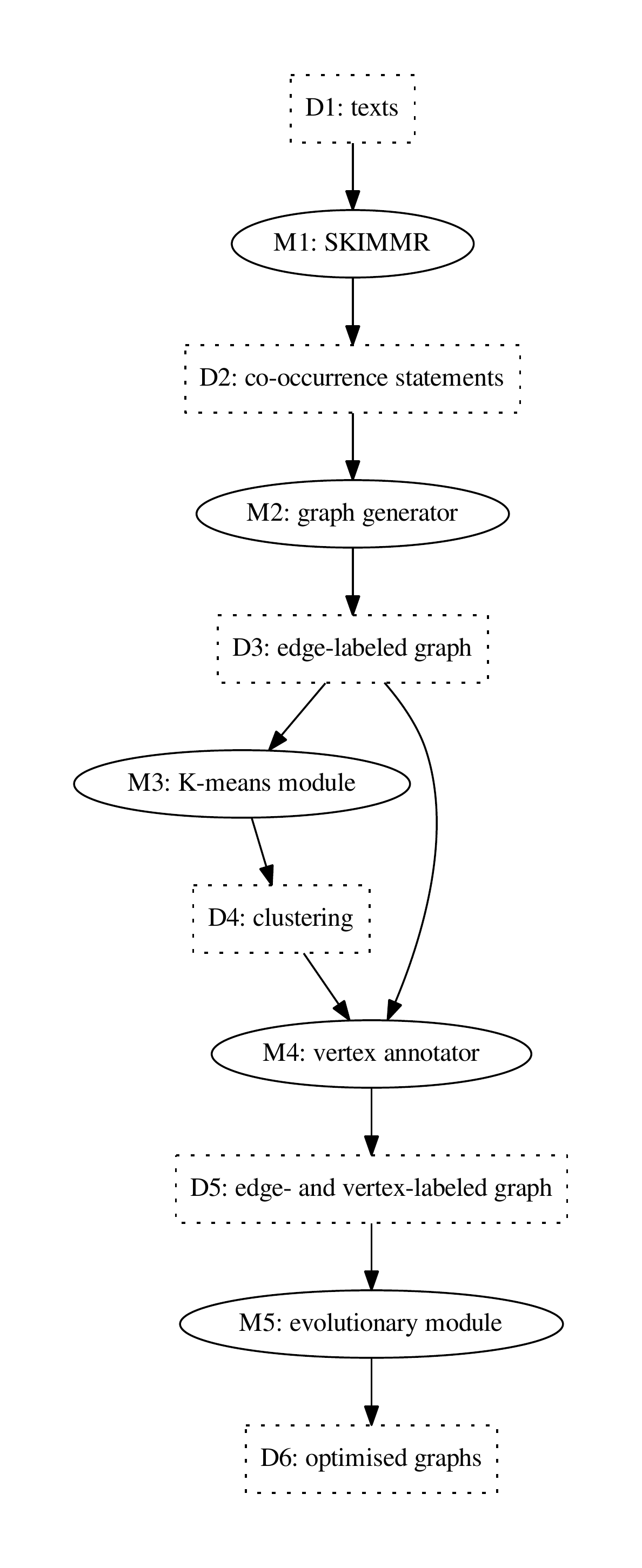}}
\caption{High-level workflow of the graph construction and 
refinement}\label{fig:workflow}
\end{figure}

Figure~\ref{fig:workflow} presents the high-level overview of the graph 
extraction and refinement process.
First we use our SKIMMR tool~\cite{novacek2014peerj} to extract basic 
co-occurrence statements from the input texts. The statements are in the 
shape of tuples $(t_1,t_2,w_d,T)$, where $t_1,t_2$ are two terms that 
co-occur in an input text $T$ and $w_d$ is the weight of the co-occurrence 
based on the sentence distances of the terms within $T$.

In the next step (M2 in Figure~\ref{fig:workflow}), we: 
\begin{enumerate}
  \item Use the basic statements to compute corpus-wide co-occurrence weights 
  using normalised point-wise mutual information.
  \item Encode the terms in the statements using integer identifiers (to 
  optimise the memory usage in the consequent steps).
  \item Build a fulltext index upon the lexical vertex labels for accessing
  them during the evaluation (this mitigates the impact of spelling 
  alternatives and other irregularities in the automatically extracted names).
  \item Initialise an undirected edge-labeled universe graph $U$ with edges 
  constructed from the corpus-wide statements. The graph can possibly be 
  limited to edges with normalised point-wise mutual information weights above 
  a pre-defined threshold.
  \item Construct a context vector space for the $U$ vertices based on their 
  neighbors and corresponding edge weights.
  \item Use the vector space to compute the Euclidean distances between the 
  vertices.
\end{enumerate}

Steps M3 and M4 in Figure~\ref{fig:workflow} perform the K-means clustering
of the universe graph $U$ in order to provide a vertex labeling $\gamma$ that
associates each vertex with cluster(s) it belongs to (see 
Section~\ref{sec:evaluation.experiments.algorithm_settings} for details on the 
K-means settings in the particular experiments we conducted). At this moment, 
everything is ready for optimising $U$ according to the hypothesis virtue 
measures of its sub-graphs.

\subsubsection{Evolutionary Graph 
Refinement}\label{sec:evaluation.evolution.graph_refinement}

\begin{figure}
\center
\scalebox{0.5}{\includegraphics{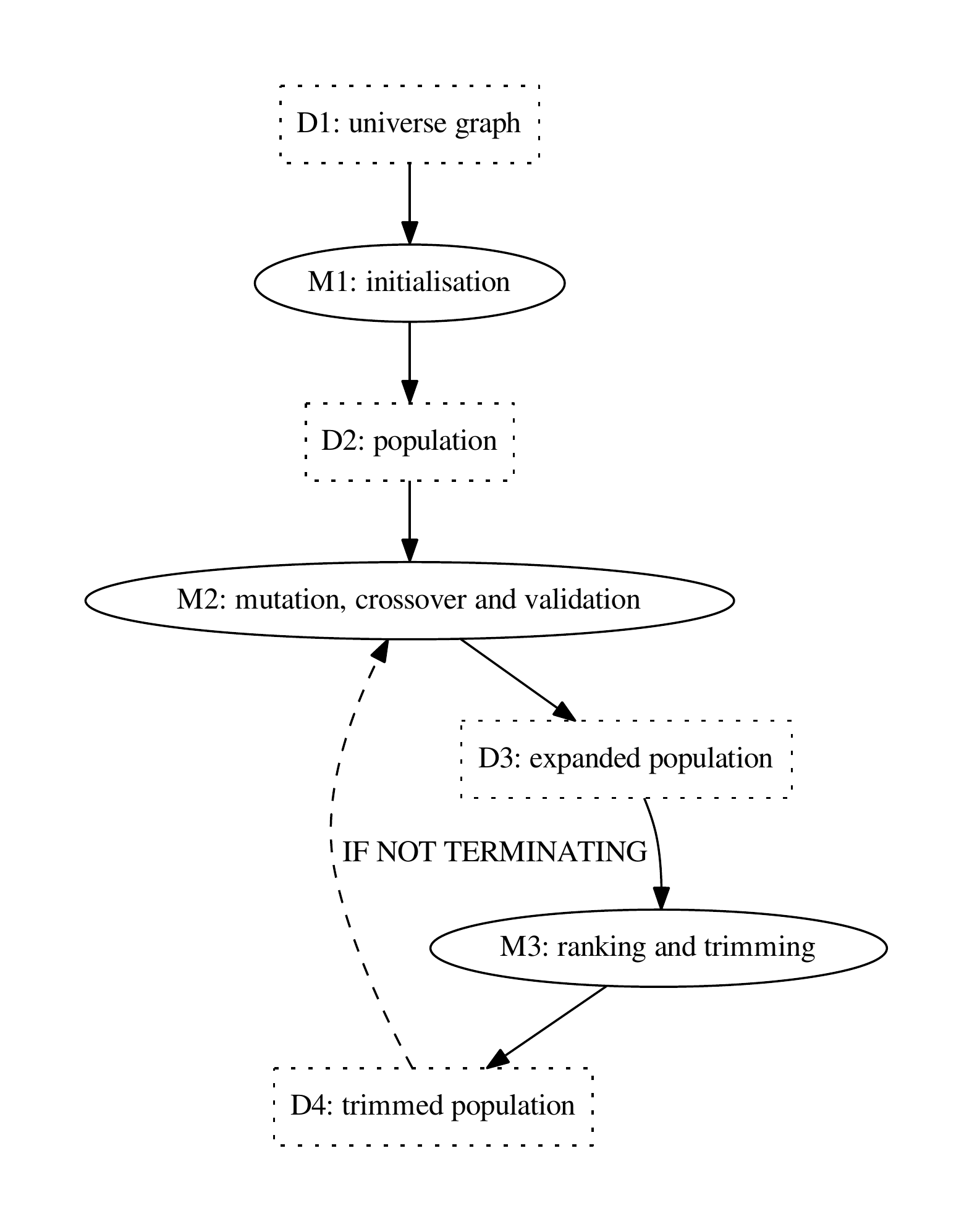}}
\caption{Detailed workflow of the evolutionary 
refinement}\label{fig:workflow_evolution}
\end{figure}

The optimisation step in Figure~\ref{fig:workflow} is performed using a 
genetic algorithm~\cite{evolcomp}. Its detailed workflow is presented in 
Figure~\ref{fig:workflow_evolution}.
The genetic algorithm has the following configurable parameters:
\begin{inparaenum}[1.]
  \item mutation and mating probabilities $p_m, p_c$ defining how likely it
  is for an individual in a population to mutate and mate (\ie engage in a 
  crossover with another individual);
  \item number $k_m$ defining how many times an individual can attempt to mate
  in a generation;
  \item maximum number of generations $N_G$;
  \item rate $\rho_p$ of the standard deviation of the population size -- it 
  sets the size of the population $P_i$ to $|P_i| = gauss(|P_{i-1}|,
  \rho_p|P_{i-1}|)$ where $gauss(\mu,\sigma)$ returns a random number from the 
  normal distribution with mean $\mu$ and standard deviation $\sigma$, 
  truncated to integer;
  \item the mean and standard deviation $\mu_i, \sigma_i$ for determining the
  sizes of the individuals in the initial population. 
\end{inparaenum}
For specific values of the parameters and discussion of their influence on the 
evolution process in our experiments, see 
Section~\ref{sec:evaluation.experiments.algorithm_settings}. 

The population is initialised (step M1 in Figure~\ref{fig:workflow_evolution}) 
by a repetitive random selection of possibly overlapping stars of size 
$gauss(\mu_i,\sigma_i)$ from the graph $U$. Stars consist of one ``hub'' vertex
and a set of vertices ``fanning out'' of the hub via immediate edges. They are 
a specific type of sub-graphs that can be used as atomic graph construction 
blocks~\cite{e14030559} and thus they are fitting for the purpose of population
initialisation. 

Step M2 in Figure~\ref{fig:workflow_evolution} consists of applying the 
evolutionary operators on the population and consequent validation of the
newly added individuals which discards disconnected ones. The mutation deletes
or adds an edge from/to the individual graph with equal probabilities. The 
crossover combines two parents by randomly selecting half of the edges from 
each parent and combining them in a new individual. All existing edge labels 
are copied in the process of creating new individuals.

Step M3 in Figure~\ref{fig:workflow_evolution} is essential for the 
optimisation -- it computes the hypothesis virtue measures of each individual
in the expanded population and then ranks the population according to the
combined ranking $\succ$ introduced in Section~\ref{sec:measures.combination}.
The population is then trimmed to a random size based on the previous 
population size (computed using the $\rho_p$ parameter). 

Steps M2 and M3 are repeated until a termination condition is met. This can 
either be reaching a pre-defined number of generations $N_G$, or achieving
some sort of population convergence.

\subsection{Description of the Experiments}\label{sec:evaluation.experiments}

For the evaluation of our approach 
we chose two standard scenarios in literature-based
discovery based on the works~\cite{swanson1986fish,swanson1987migraine}. 
Details on the corresponding data sets and experiments we performed using them 
are described in the following sections.

\subsubsection{Data 
Acquisition}\label{sec:evaluation.experiments.data_acquisition}

We used two data sets in the experimental evaluation, both of which address 
discovery of connections between previously isolated concepts (and 
corresponding bodies of literature). One data set is based 
on~\cite{swanson1986fish} that explores the relationship between fish oil and
Raynaud's syndrome. The other data set is based on similar study of previously
neglected connections between migraine and 
magnesium~\cite{swanson1987migraine}. We refer to these two data sets and
corresponding experiments as to $T_{R}, T_{M}$, respectively. 

The initial corpora of texts for the $T_{R}, T_{M}$ experiments were obtained
from PubMed via queries compiled according to the specifications given 
in~\cite{swanson1986fish,swanson1987migraine}. Each of these works defines 
source and target terms $t_s, t_t$ together with a set $I_c$ of intermediate 
terms that connect them. A query for the PubMed abstracts corresponding to 
specific $t_s, t_t, I_c$ is compiled as a disjunction of atomic conjunctions
$$
\bigvee_{t \in \{t_s, t_t\}, t_c \in I_c} (t \wedge t_c)
$$

The particular queries we used for obtaining the $T_{R}, T_{M}$ corpora
were
\begin{quote}
\small\tt
("raynaud" AND "blood") OR ("raynaud" AND "viscosity") OR \\ 
("raynaud" AND "platelet") OR ("raynaud" AND "vascular") OR \\
("raynaud" AND "reactivity") OR ("fish oil" AND "blood") OR \\
("fish oil" AND "viscosity") OR ("fish oil" AND "platelet") OR \\
("fish oil" AND "vascular") OR ("fish oil" AND "reactivity")
\end{quote}
and
\begin{quote}
\small\tt
("migraine" AND "vasospasm") OR ("migraine" AND "spreading depres\-sion") OR 
("migraine" AND "vascular reactivity") OR ("migraine" AND "depolarization") OR
("migraine" AND "epilepsy") OR ("migraine" AND "inflammation") OR
("migraine" AND "pro\-sta\-glan\-dins") OR \\ ("migraine" AND "platelet 
aggregation") OR 
("migraine" AND "sero\-to\-nin") OR ("migraine" AND "brain anoxia") OR 
("migraine" AND "cal\-cium channel blockers") OR ("magnesium" AND "vasospasm") 
OR ("mag\-ne\-sium" AND "spreading depression") OR ("magnesium" AND "vascular 
reactivity") OR ("magnesium" AND "depolarization") OR ("magnesium" AND 
"epilepsy") OR ("magnesium" AND "inflammation") OR ("magnesium" AND 
"prostaglandins") OR ("magnesium" AND "platelet aggregation") OR 
("magnesium" AND "serotonin") OR ("magnesium" AND "bra\-in a\-no\-xia") OR 
("magnesium" AND "calcium channel blockers")
\end{quote}
respectively. Note that the while the $T_{M}$ query exactly corresponds 
to the terms given in~\cite{swanson1987migraine}, the $T_{R}$ query is 
relaxed to sub-terms as the exact query only yields very few abstracts. The
PubMed search was limited to articles indexed until November, 1985 and 
August, 1987 for $T_{R}, T_{M}$, respectively, so that we can compare 
ourselves to the findings of the original works which have served as a 
{\it de facto} gold standard in the literature-based discovery 
field~\cite{Cameron2015}. 

The characteristics of the $T_{R}, T_{M}$ corpora are summarised in 
Table~\ref{tab:corpora_stats}.
\begin{table}[ht]
\footnotesize
\center
\begin{tabular}{|l||c|c|c|}
\hline
Corpus   & \# of abstracts & \# of tokens & \# of base statements \\ 
\hline\hline
$T_{R}$ & 1,406           & 90,427        & 407,154              \\ 
\hline
$T_{M}$ & 3,611           & 319,810       & 1,534,685            \\ 
\hline
\end{tabular}
\caption{Basic statistics of the corpora}\label{tab:corpora_stats}
\end{table}
Number of tokens is a sum of the word-length of the documents in the corpus 
and number of base statements is the number of the base co-occurrence 
statements the SKIMMR tool extracted from the corpus.

\subsubsection{Graph 
Extraction}\label{sec:evaluation.experiments.graph_extraction}

To generate knowledge graphs from the text corpora, we use the approach 
introduced in Section~\ref{sec:evaluation.evolution}. We construct the
experimental graphs using only co-occurrence statements with above-average 
positive normalised point-wise mutual information scores. This filters out 
statements with comparatively low co-occurrence weight. We use the general 
SKIMMR version that extracts entities based on shallow parsing rather than 
domain-specific models (see \url{https://github.com/vitnov/SKIMMR} for 
details). This is to demonstrate the generality of our work -- if we show 
that our approach can deliver good results even in quite a specific domain 
using basic and universally applicable initial text mining, it indicates that 
it is likely to perform similarly well in any other domain.

\begin{figure}
\center
\scalebox{0.5}{\includegraphics{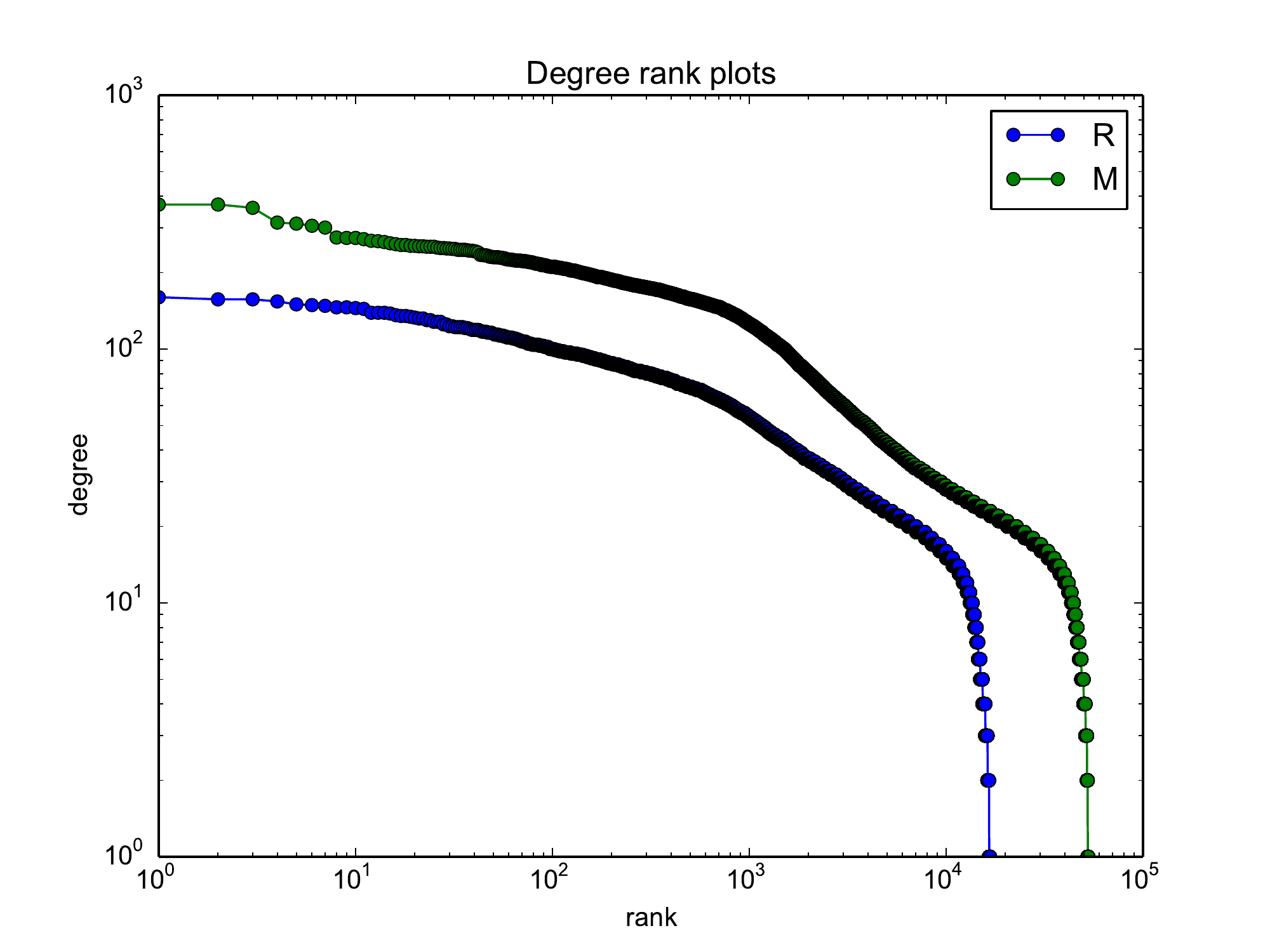}}
\caption{Degree distributions for the experimental graphs}\label{fig:degdist}
\end{figure}

The characteristics of the extracted graphs are provided in 
Tables~\ref{tab:graph_bstats} and~\ref{tab:graph_cstats}.
\begin{table}[ht]
\footnotesize
\center
\begin{tabular}{|l||c|c|c|c|c|c|c|}
\hline
Graph    & $|V_G|$       & $|E_G|$       & $dn_G$   & $|C_G|$ & $|c_G^{max}|$ & 
           $|c_G^{avg}|$ & $|c_G^{med}|$ \\ 
\hline\hline
$T_{R}$ & 16,714 & 181,140 & 1.297e-3 & 80  & 16,497 & 208.925 & 2 \\
\hline
$T_{M}$ & 52,681 & 635,705 & 4.581e-4 & 110 & 52,373 & 478.918 & 2 \\
\hline
\end{tabular}
\caption{Basic characteristics of the experimental 
graphs}\label{tab:graph_bstats}
\end{table}
The basic characteristics $|V_G|, |E_G|, dn_G, |C_G|, |c_G^{max}|,
|c_G^{avg}|,|c_G^{med}|$ in Table~\ref{tab:graph_bstats} are the number of 
vertices, number of edges, graph density, number of connected components, 
maximum, average and median component size in vertices, respectively. The 
component-wise characteristics in Table~\ref{tab:graph_cstats} are computed 
as a weighed arithmetic mean across all the components where the weight is the
component size in vertices. 
\begin{table}[ht]
\footnotesize
\center
\begin{tabular}{|l||c|c|c|c|c|}
\hline
Graph    & $rd_G$ & $dm_G$ & $tr_G$  & $asp_G$ & $asp_G^{\delta}$ \\ 
\hline\hline
$T_{R}$ & 5.935  & 8.897  & 0.397   & 4.045   & 6.956     \\
\hline
$T_{M}$ & 5.971  & 8.954  & 0.259   & 3.964   & 7.702     \\
\hline
\end{tabular}
\caption{Component-wise characteristics of the experimental 
graphs}\label{tab:graph_cstats}
\end{table}
The characteristics $rd_G, dm_G$ are the graph radius and diameter
(minimum and maximum eccentricity, respectively, where eccentricity of a vertex
is its maximum distance to other vertices). The $tr_G$ characteristics is 
transitivity -- the fraction of all possible triangles reflecting the tendency 
of vertices in the graph to cluster together~\cite{sna}. The characteristics
$asp_G, asp_G^{\delta}$ are average shortest path lengths in terms of edges 
and the distance labeling, respectively. Additional characteristics of the 
graph is the degree distribution depicted in Figure~\ref{fig:degdist} (the 
plot is log-scaled in both x- and y-axis).

The extracted graphs both have one large connected component comprising most
of the vertices, complemented by other trivial components mostly consisting of 
one edge. The largest components exhibit so called ``small-world'' 
property~\cite{watts1998cds} -- despite of being quite large and having small 
density, they have relatively small diameters and average shortest paths. This
observation is supported by two additional facts. The graphs have relatively 
high transitivity, \ie high tendency of vertices to cluster together which is 
typical for complex small-world networks~\cite{sna}. Also, the vertex degree 
distribution approximately follows the power law as shown in 
Figure~\ref{fig:degdist}, which is characteristic for scale-free 
networks~\cite{onnela2007structure}. This means that the extracted graphs
have relatively densely connected structure with many claims involving
frequently repeated concepts, which is largely caused by highly (co)occurring 
terms. This is perhaps not ideal for making discoveries about as many 
previously disconnected phenomena as possible, and we later show how our 
approach can remedy this problem.

\subsubsection{Settings of the Clustering and Evolutionary 
Refinement Algorithms}\label{sec:evaluation.experiments.algorithm_settings}

For the clustering, we use the K-means module of the {\it scikit-learn} 
package~\cite{scikit-learn}. As the algorithm's scalability to large numbers 
of samples and features is limited by available memory, we partition the set 
of context vectors corresponding to the universe graph vertices to buckets of 
size $2,000$ and then run the K-means algorithm on them with the parameter $K$ 
set to $40$. The partitioning is done by incremental random selection of $50$ 
seed vectors from the unpartitioned set, computing their centroid and then 
filling the partition with the seeds plus up to $1,950$ unpartitioned vectors 
closest to the centroid. We have experimented with different settings of every 
parameter, however, we found out that the resulting distributions of vectors 
into clusters are practically invariant to the settings, with mean and median 
cluster sizes converging to the same values no matter what the settings were.

The parameters of the evolutionary refinement were
$$
p_m = 0.05, p_c = 0.75, k_m = 5, N_G = 50, \rho_p = 0.05, \mu_i = 100, 
\sigma_i = 80.
$$
The initial individual size parameters are only reflected in rare extreme cases
as the size of the random stars is much more dependent on the data set 
structure in practice. For the other parameters, we applied values typically 
used by model approaches presented in the evolutionary computing 
literature~\cite{evolcomp}. The number of generations has been set well above
a threshold after which the performance of the corresponding populations starts
to oscillate around similar evaluation scores (see 
Section~\ref{sec:evaluation.results.selection} for details). 

The evolutionary refinement with these parameters took 56m and 6h36m for the 
$T_R, T_M$ experiments, respectively, using a 2010 make laptop with 4-core CPU,
8GB RAM and Ubuntu Linux 14.04 OS. The virtue measures (the most 
demanding part) were computed using six parallel processes. The number of
processes can be easily adjusted to the computing power available, which 
facilitates vertical scalability of the refinement. Horizontal scalability
is planned for future versions of the prototype and consists of using a 
distributed processing library instead of the native Python multiprocessing 
module. 

\subsubsection{Evaluation 
Methodology}\label{sec:evaluation.experiments.evaluation_methodology}

We use several evaluation methods. Part of them is based on a recent 
work~\cite{Cameron2015} which defines evidence-based and literature 
frequency-based evaluation measures within the {\it de facto} standard 
literature-based discovery scenarios elaborated in~\cite{swanson1986fish,
swanson1987migraine}. The additional benefit of using~\cite{Cameron2015} as a 
primary reference for the evaluation is that the authors compared results of 
several representative approaches to literature-based discovery. Thus we can
interpret our results within a broader context of the whole field. In 
addition to the measures defined in~\cite{Cameron2015}, 
we perform qualitative evaluation 
of the actual contents of the results and compare ourselves to related state 
of the art where applicable.

The evidence-based evaluation measures the capability of an approach to 
re-discover the intermediate concepts linking the source and target in the
corpus as per discoveries made by human experts. It also measures the 
importance the approach associates with the re-discovery. For an intermediate 
$t_c$, the absolute evidence-based evaluation measure directly corresponding 
to~\cite{Cameron2015} is defined as
$$
evd(t_c) = \min_{G \in \mathcal{G}_c}(rnk(G)),
$$
where $\mathcal{G}_c = \{G|t_s, t_t, t_c \in V_G \wedge \exists p \in \Pi(G) . 
p = (t_s, \dots, t_c, \dots, t_t)\}$ is a set of solution graphs that contain 
the source and target terms $t_s,t_t$ linked by the intermediate 
$t_c$\footnote{Note that for mapping terms to vertices in the resulting 
knowledge graphs, we use the fulltext index computed upon the lexical 
expressions corresponding to the graph vertices. This is done when generating 
the universe graph, see 
Section~\ref{sec:evaluation.evolution.graph_construction} for details. 
To get all term manifestations in our automatically extracted knowledge 
graphs, we look up the term of interest in the index and then manually prune 
the results to get all alternatives that refer to the corresponding concept.}. 
The function $rnk: \mathcal{G} \rightarrow \mathbb{N}$ is a ranking of all 
solution graphs $\mathcal{G} = \{G|t_s, t_t \in V_G\}$ from the most to the 
least relevant where the relevance is determined by the specific approach 
being evaluated.

We construct the sets of ranked solution graphs from the set of individuals in 
a selected refined generation by:
\begin{inparaenum}[1.]
  \item Creating a union graph from all population individuals.
  \item Generating a set of paths between the source and target term vertices
  that also contain an intermediate vertex.
  \item Ranking the paths using their hypothesis virtue measures, \ie the 
  $\succ$ relation, with the population union graph as a universe.
\end{inparaenum}
The step 2. can either compute all simple paths or all shortest paths. 
In our experiments, we use the latter option due to 
tractability issues. The conception of paths as solution graphs represents 
another design choice consistent with the previous definitions -- a path
linking certain concepts is the simplest way of claiming (and potentially also
explaining) something about them.

In addition to the absolute $evd$ score, we compute the overall relative 
importance of an intermediate term $t_c$. 
This measure 
is defined as a mean relative inverse rank of the graphs that contain $t_c$
among all solutions, \ie
$$
evd^r(t_c) = \frac{1}{|\mathcal{G}_c|}\sum_{G \in \mathcal{G}} 
\frac{|\mathcal{G}| - rnk(G) + 1}{|\mathcal{G}|}.
$$
It effectively measures the average relative relevance of the hypotheses 
linking the source and target terms via $t_c$ -- the more often the link is 
discovered in high-ranking graphs, the higher the measure. 

The second evaluation method proposed in~\cite{Cameron2015} measures the
frequency of the discovered claims in the scientific literature. Similarly to 
our definition, a path in the result graph is considered a claim 
in~\cite{Cameron2015}. The literature frequency can be used to define a 
measure of solution rarity as
$$
rar(\mathcal{G}) = 
\frac{1}{|\pi_s(\mathcal{G}_I)|} \sum_{p \in \pi_s(\mathcal{G}_I)} 
f_{pm}(Q_A(p)),
$$
where $\mathcal{G}_I = \{G|G \in \mathcal{G} \wedge \exists t_c \in I_c, p \in 
\Pi(G).p = (t_s,\dots,t_c,\dots,t_t)\}$ is a set of solutions that contain an
intermediate term, $\pi_s(\mathcal{G}_I) = \bigcup_{G \in \mathcal{G}}\pi_s(G)$
is the union of shortest paths 
taken across $\mathcal{G}_I$, and $f_{pm}$ is the number of results returned by
PubMed for an association query $Q_A(p)$. The query for 
a path $(p_1,p_2,\dots,p_{|p|})$ corresponds to the conjunction 
$\bigwedge_{t \in p} t$ of all terms in the path (with a publication time 
window limited according to the corresponding experimental corpus). For 
instance, the path {\it (fish oil, platelet aggregation, Raynaud's syndrome)} 
corresponds to the PubMed query {\tt "fish oil" AND "platelet aggregation" AND 
"Raynaud's syndrome" AND ("0001/01/01"[PDAT] : "1985/11/30"[PDAT])} in the 
$T_R$ experiment. Finally, the rarity measure can be straightforwardly used 
for defining an interestingness measure~\cite{Cameron2015} as a normalised 
inverse of the rarity
$$
int(\mathcal{G}) = \frac{1}{1+rar(\mathcal{G})}.
$$

The qualitative evaluation of the results is based on the sets of topics
covered by the particular solutions. A topic is informally defined by 
potentially relevant terms that lay on a path between source and target 
concepts in a solution. Potentially relevant terms are those that refer to
non-trivial concepts that may elucidate the meaning of the particular path. 
Using the notion of topics, we define the measures of topical density,
relative topical relevance and relative topical novelty, respectively, as
$$
top_d(\mathcal{G}) = \frac{|T_{unq}(\mathcal{G}_I)|}{|T_{all}(\mathcal{G}_I)|},
\;
top_r(\mathcal{G}) = \frac{|T_{rel}(\mathcal{G}_I)|}{|T_{unq}(\mathcal{G}_I)|},
\;
top_n(\mathcal{G}) = \frac{|T_{nvl}(\mathcal{G}_I)|}{|T_{rel}(\mathcal{G}_I)|}
$$
for a set $\mathcal{G}$ of all solution graphs. The sets 
$T_{unq}(\mathcal{G}_I), T_{all}(\mathcal{G}_I), T_{rel}(\mathcal{G}_I), 
T_{nvl}(\mathcal{G}_I)$ are sets of unique, all, relevant and novel topics 
covered by the solution graphs in $\mathcal{G}$ that contain an intermediate 
term. 

The relevance of topics is determined by a review of the available scientific 
literature. This tells us whether or not a given set of terms can provide a 
meaningful and non-trivial explanation of the connection between the source and
target terms. More specifically, a topic is considered relevant if and only 
if the following conditions are met simultaneously:
\begin{inparaenum}[1.]
  \item The terms in the topic refer to features of a biomedically relevant
  relationship that can be traced in literature. 
  \item The relationship is associated with the corresponding target, 
  source and intermediate terms.
  \item The relationship is not trivial -- it has to be a supported by genuine
  discoveries presented in literature, not statements of obvious merely 
  occurring in articles.
\end{inparaenum}

A novel topic is one that is relevant and not covered by any single published 
work in its whole. This can be determined using a publication search engine 
such as PubMed, where we can check the number of results of a conjunctive 
query involving all terms in the corresponding claim path. If the number of
results is zero, then the topic is unique.

We compute the $top_d,\;top_r,\;top_n$ scores for the initially extracted and 
refined graphs in both experiments, focusing on solutions involving 
corresponding source, target and intermediate terms. 
Whenever applicable, we compare the relevant topics we generated with the 
topics (re)discovered by related approaches. 

\subsection{Results and Discussion}\label{sec:evaluation.results}

We split this section into three parts -- first we explain the process of
selection of the refined graphs to be evaluated, then we analyse properties of 
the selected graphs, and finally we discuss the results of the evaluation. 

\subsubsection{Selection of the Refined 
Graphs}\label{sec:evaluation.results.selection}

\begin{figure}
\center
\scalebox{0.5}{\includegraphics{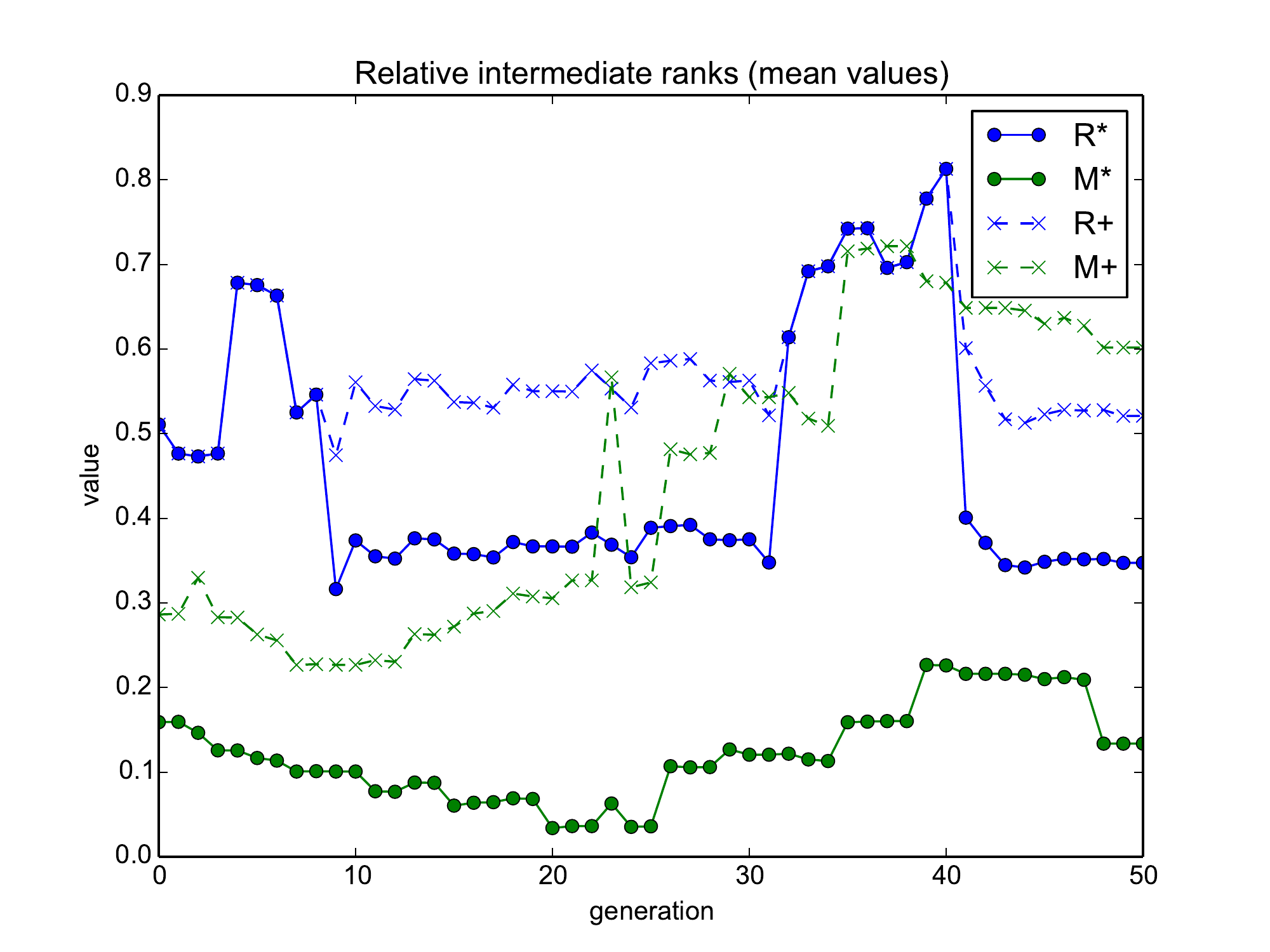}
}
\caption{Mean relative intermediate ranks through 
generations}\label{fig:evd_evol}
\end{figure}

Before analysing the actual results of the evolutionary refinement, we have to
select the generation we will focus on. A natural criterion for that is the
performance of generations in terms of the evaluation measures. The relative
ranking of intermediate concepts (\ie the $evd^r$ measure) is best suited for 
this task as it tells us to which extent the generations tend to ``consider'' 
the intermediate connections important. Figure~\ref{fig:evd_evol} shows how 
the mean $evd^r$ values for all intermediates evolve throughout the 
generations for each experiment.
The blue and green lines represent the $T_R, T_M$ experiments, respectively.
The full lines correspond to mean values taken across all intermediate terms
(also marked by the ``star'' character in the plot legend). The dashed lines 
are for mean values omitting intermediates that are not present in the given 
generation (marked by the ``plus'' character in the legend).

For the $T_R$ experiment, the generation 40 clearly performs best as it 
contains solution graphs for each intermediate term and their mean
relative ranking is very high (within the top 20\% of solutions). For the
$T_M$ experiment, the situation is less clear. 
The best generation in terms of mean across all intermediate
terms is number 39, however, if one takes only the present intermediates into
account, the generations 35-38 all perform better. Yet we decided to further
analyse the generation 39 as it covers three intermediates, while the 
generations 35-38 only cover two. From here on, we refer to the selected 
generations by the $T_R^{40},T_M^{39}$ expressions, respectively.

\begin{figure}
\center
\scalebox{0.5}{
\includegraphics{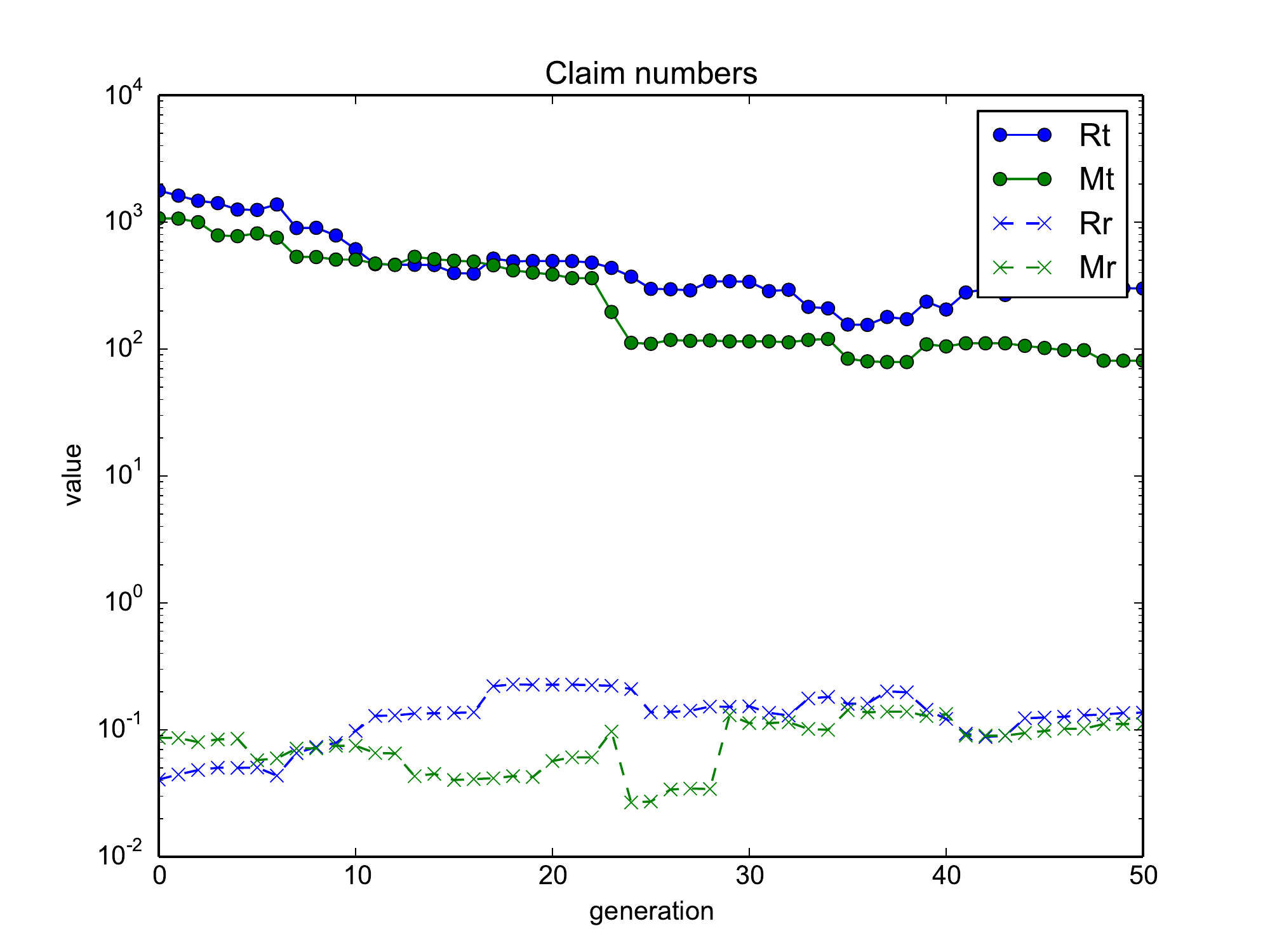}
}
\caption{Claim numbers through generations}\label{fig:claim_number_evol}
\end{figure}

Further support for selecting the generation to be analysed can be drawn from 
the numbers of claims containing the source and target claims, and the numbers 
of such claims that also contain an intermediate term. The evolution of these 
values is depicted in Figure~\ref{fig:claim_number_evol}.
Note that the figure's y-axis is log-scaled due to different orders of 
magnitude of the displayed values. Similarly to the previous figure, 
the blue and green lines represent the $T_R, T_M$ experiments, respectively.
The full lines correspond to the total number of claims containing the source 
and target term in a given generation (also marked by the ``t'' character in 
the plot legend). The dashed lines represent the fraction of the claims that 
also contain an intermediate term (marked by the ``r'' character).

The total number of relevant claims is steadily decreasing up until 
approximately 20-25th generation and then starts to oscillate. For the 
relative number of solutions with intermediates, similar trend can be seen 
after the 40-th generation. This can be interpreted as an indication that the 
generations are structurally stabilised then, at least for the evaluation data 
we work with. 

\subsubsection{Properties of the Refined 
Graphs}\label{sec:evaluation.results.properties}

\begin{figure}
\center
\scalebox{0.5}{\includegraphics{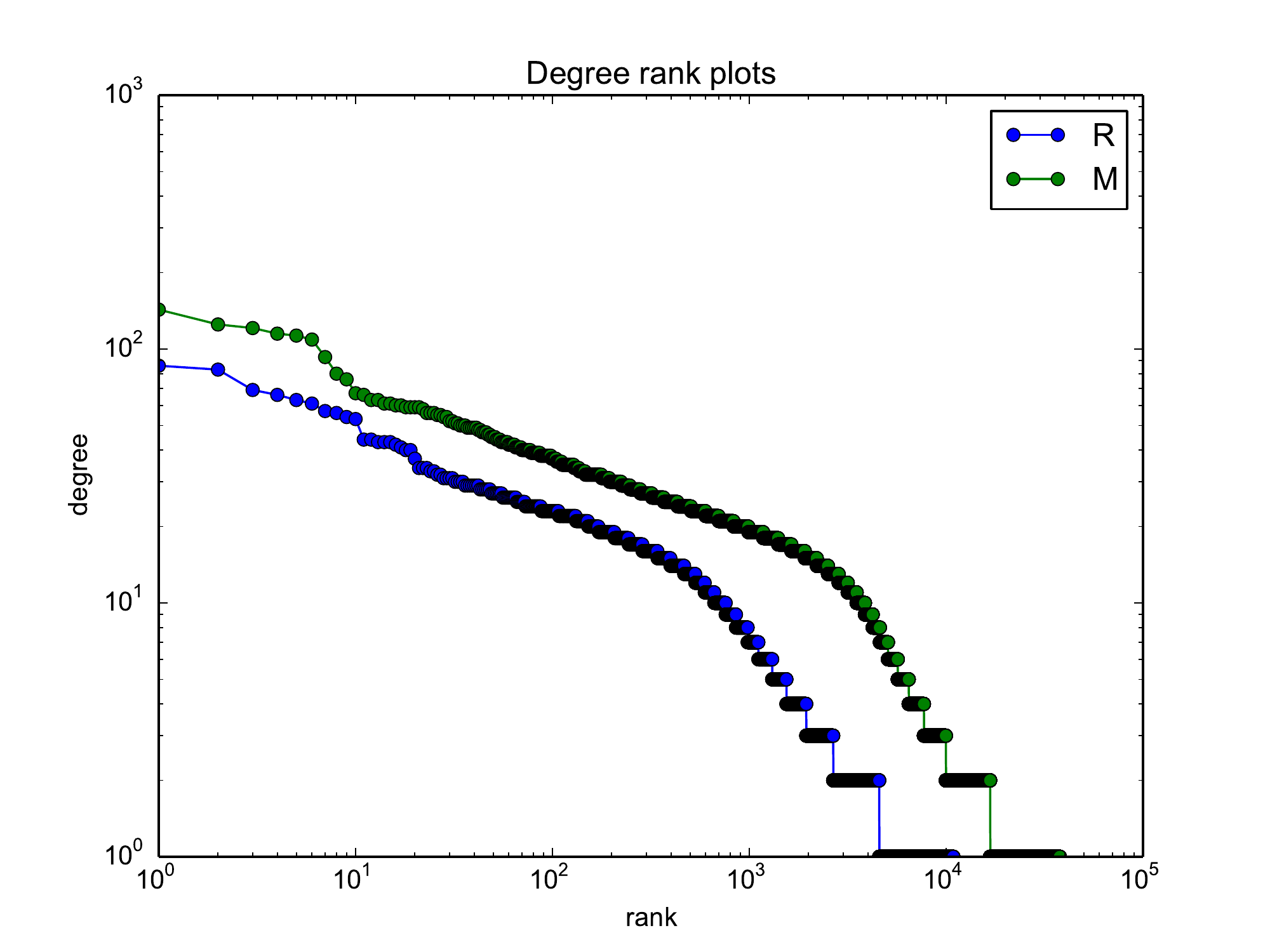}}
\caption{Degree distributions for the evolved experimental 
graphs}\label{fig:degdist_evolved}
\end{figure}

Before we proceed with discussing the results, let us have a look at the 
characteristics of the knowledge graphs corresponding to the generations we 
selected for evaluation. Tables~\ref{tab:evolved_graph_bstats} 
and~\ref{tab:evolved_graph_cstats} present the same type of data like the 
tables in Section~\ref{sec:evaluation.experiments.graph_extraction}. 
The extra rows with the $\Delta$ prefixes show the relative difference between 
the refined and initial graphs. 
\begin{table}[ht]
\footnotesize
\center
\begin{tabular}{|l||c|c|c|c|c|c|c|}
\hline
Graph    & $|V_G|$       & $|E_G|$       & $dn_G$   & $|C_G|$ & $|c_G^{max}|$ & 
           $|c_G^{avg}|$ & $|c_G^{med}|$ \\ 
\hline\hline
$T_{R}^{40}$   & 10,879  & 15,940  & 2.694e-4 & 81  & 10,670 & 134.309 & 2 \\
\hline
$\Delta T_{R}$ & \it 0.651   & \it 0.088   & \it 0.208    & \it 1.013  & \it 0.647  & \it 0.643   & \it 1 \\
\hline
$T_{M}^{39}$   & 37,782  & 65,263  & 9.144e-5 & 129 & 37,431 & 292.884 & 2 \\
\hline
$\Delta T_{M}$ & \it 0.717   & \it 0.103   & \it 0.2      & \it 1.173  & \it 0.715  & \it 0.612   & \it 1 \\
\hline
\end{tabular}
\caption{Basic characteristics of the evolved experimental 
graphs}\label{tab:evolved_graph_bstats}
\end{table}
The columns represent exactly the same measures as in the tables in
Section~\ref{sec:evaluation.experiments.graph_extraction} -- number of 
vertices, number of edges, graph density, number of connected components, 
maximum, average and median component size in nodes ($|V_G|, |E_G|, dn_G, 
|C_G|, |c_G^{max}|, |c_G^{avg}|,|c_G^{med}|$), and the graph radius, diameter,
transitivity and average shortest path lengths in terms of edges and the 
distance labeling ($rd_G, dm_G, tr_G, asp_G, asp_G^{\delta}$).
\begin{table}[ht]
\footnotesize
\center
\begin{tabular}{|l||c|c|c|c|c|}
\hline
Graph           & $rd_G$ & $dm_G$ & $tr_G$  & $asp_G$ & $asp_G^{\delta}$ \\ 
\hline\hline
$T_{R}^{40}$   & 8.847  & 14.743 & 0.015   & 6.722   & 12.309    \\
\hline
$\Delta T_{R}$ & \it 1.491  & \it 1.657  & \it 0.038   & \it 1.662   & \it 1.77 \\
\hline
$T_{M}^{39}$   & 7.936  & 13.885 & 0.014   & 6.349   & 13.721    \\
\hline
$\Delta T_{M}$ & \it 1.329 & \it 1.551  & \it 0.054   & \it 1.602   & \it 1.781\\
\hline
\end{tabular}
\caption{Component-wise characteristics of the evolved experimental 
graphs}\label{tab:evolved_graph_cstats}
\end{table}
Figure~\ref{fig:degdist_evolved} 
contains plots of the degree distribution in the refined graphs.

The refined graphs still contain about 65\% and 72\% of the original vertices
for the $T_R, T_M$ experiments, respectively, however, the edges are much more 
pruned, to about 9\% and 10\%, respectively. The graph density is thus lower, 
too (at about 20\%). The numbers of connected components do not change much. 
This is only to be expected given the nature of the population preparation and 
the tendency of the evolution process to preserve connectedness. The sizes of 
the components are more or less proportional to the reduction of the vertex 
number. 

What is more interesting 
are the 
component-wise characteristics of the refined graphs summarised in 
Table~\ref{tab:evolved_graph_cstats}. The radius, diameter and average 
shortest path lengths are all increased by up to 78\% and no less than 32\%, 
despite of the graphs getting smaller. The clustering coefficient decreases 
quite radically -- to about 3.8\% and 5.4\% of the original value for the 
$T_R,T_M$ experiments, respectively. The vertex degree distribution still 
approximately follows the power law, however, the curve is not as steep as for 
the original graphs. These combined characteristics indicate that the refined 
graphs exhibit the small world property to much lower extent than the 
original ones. This means that they are structurally more evenly organised and 
tend to have less vertices or vertex groups that connect large portions of the 
graph through very few edges. A possible consequence of this fact is lower 
redundancy and higher rate of non-obvious connections in the refined graphs.
Indeed, the analysis of the data \wrt the standard literature-based discovery 
application scenarios confirms this, as we show in the next section.

\subsubsection{Performance of the Refined 
Graphs}\label{sec:evaluation.results.evaluation}

In this section, we first discuss the performance of our experiments \wrt the 
quantitative measures used by related approaches.
This is then followed by 
qualitative analysis of the knowledge graphs we generated.

Table~\ref{tab:evaluation_evidence.rf} lists the values of the $evd$ measure
for the $T_{R}$ data set. Our approach (the N column) is compared to the 
works~\cite{Cameron2015,srinivasan2004text,weeber2001using,
gordon1996toward,hristovski2006exploiting} in columns C, S, W, G, H, 
respectively. For our approach, we list both $evd, evd^r$ values, while for
the others, only $evd$ is present as they do not consider $evd^r$. We also
provide $|G_c|$, \ie the number of solution graphs with intermediates.
\begin{table}[ht]
\footnotesize
\center
\begin{tabular}{|l||c|c|c|c|c|c|c|c|}
\hline
Intermediate & \multicolumn{3}{|c|}{N}     & C~\cite{Cameron2015}
             & S~\cite{srinivasan2004text} & W~\cite{weeber2001using}
             & G~\cite{gordon1996toward}   & H~\cite{hristovski2006exploiting}
\\
             & $evd$ & $evd^r$ & $|G_c|$   &
             &                             &
             &                             &
\\
\hline\hline
Blood Viscosity      & 5  & 0.98  & 1  & 15$^*$ & 2 & Y & 5  & 8 \\ 
\hline
Platelet Aggregation & 20 & 0.844 & 18 & 1      & 1 & Y & 6  & 17 \\ 
\hline
Vascular Reactivity  & 73 & 0.615 & 4  & -      & 1 & Y & 19 & -  \\
\hline
\end{tabular}
\caption{Evidence-based evaluation -- $T_{R}^{40}$ 
data}\label{tab:evaluation_evidence.rf}
\end{table}
The $evd$ numbers correspond to the best rank of the result that contains given
intermediate term. The ``-'' character means the intermediate cannot be found
in any result for that approach. If there is ``Y'', then the intermediate can
be found in the results but no ranking is provided. Finally, the results with 
``$^*$'' in the C column indicate that the intermediate can only be found 
indirectly by manually exploring a broader context of the 
result~\cite{Cameron2015}.

Our approach finds all intermediate terms which makes its performance 
equivalent to or better than the related approaches in this respect. Blood 
viscosity and platelet aggregation are placed among the top 16\% of the 
results (out of 205 in the $T_R$ experiment) while vascular reactivity is 
considered to be relatively less important intermediate. 

Table~\ref{tab:evaluation_evidence.mm} lists the same type of results as
Table~\ref{tab:evaluation_evidence.rf}, only for the $T_{M}$ experiment and
slightly different set of related works.
\begin{table}[ht]
\footnotesize
\center
\begin{tabular}{|l||c|c|c|c|c|c|c|c|}
\hline
Intermediate & \multicolumn{3}{|c|}{N}         & C~\cite{Cameron2015}
             & S~\cite{srinivasan2004text}     & W~\cite{weeber2001using}
             & B~\cite{blake2002automatically} & G~\cite{gordon1996toward}
\\
             & $evd$ & $evd^r$ & $|G_c|$   &
             &                             &
             &                             &
\\
\hline\hline
Calcium Channel                 &    &       &    &       &   &   &     &   \\
Blockers                        & -  & -     & -  & 22    & 3 & Y & 10  & 1 \\ 
\hline
Epilepsy$^{39}$                 & 23 & 0.628 & 7  & 9$^*$ & - & Y & 8   & 3 \\ 
\hline
Brain Anoxia /                  &    &       &    &       &   &   &     &   \\
Hypoxia                         & -  & -     & -  & -     & 5 & - & 6   & 77 \\ 
\hline
Inflammation                    & -  & -     & -  & 3$^*$ & 2 & Y & 170 & 82 \\ 
\hline
Platelet Aggregation /          &    &       &    &       &   &   &     &   \\
Activity$^{10}$                 & 335& 0.333 & 2  & 1$^*$ & 2 & Y & 2   & 8 \\ 
\hline
Prostaglandins$^{10}$           & 352& 0.274 & 4  & 4     & 1 & Y & 42  & 27 \\ 
\hline
Serotonin                       & -  & -     & -  & 1     & 1 & Y & 5   & 1 \\ 
\hline
Cortical / Spreading            &    &       &    &       &   &   &     &   \\ 
Depression$^{39}$               & 58 & 0.468 & 3  & -     & 6 & - & 45  & - \\ 
\hline
Vascular Mechanism /            &    &       &    &       &   &   &     &   \\ 
Reactivity$^{39}$               & 7  & 0.945 & 1  & 9     & 1 & Y & 46  & 16 \\ 
\hline
\end{tabular}
\caption{Evidence-based evaluation -- $T_{M}^{39}, T_{M}^{10}$ 
data}\label{tab:evaluation_evidence.mm}
\end{table}
Note that the related works are sometimes inconsistent in the exact wording
of the intermediate terms, therefore we only focused on nine out of eleven
where we were able to clearly mash up the different alternatives of the term.

The quantitative results of our approach are sparser than in case of the
$T_R$ experiment. This has been caused mainly by the minimalistic, 
domain-agnostic approach we chose, which resulted into relatively low coverage
of the intermediate synonyms appearing in the data (the fulltext mapping could
only discover terms rather similar to the canonical intermediate form used
as a query, while many synonyms are quite dissimilar strings). All related 
approaches but one~\cite{gordon1996toward} use term expansion and mapping 
using biomedical vocabularies like MeSH, and some even use quite extensive 
manual interventions (see Section~\ref{sec:related.lbd} for details). Despite 
of these limitations, we re-discovered five out of nine intermediates. Out of 
these, only three were discovered using a mature-enough generation of the 
refined knowledge graph, though. 

For the intermediates we managed to find, we achieved results comparable to or 
better than the other approaches. For instance, three out of five related 
approaches were not able to re-discover the cortical depression intermediate
which is considered very important in~\cite{swanson1987migraine}. 

The overall results of the evidence-based evaluation 
are 
encouraging. In the $T_R$ experiment, our approach performed better 
than~\cite{Cameron2015,hristovski2006exploiting}, worse 
than~\cite{srinivasan2004text,gordon1996toward} and equally 
to~\cite{weeber2001using}. In the $T_M$ experiment, we 
bettered~\cite{Cameron2015,weeber2001using,gordon1996toward} 
while~\cite{srinivasan2004text,blake2002automatically} outperformed 
us\footnote{Note that direct comparison of the ranking results is conceptually 
difficult since the approaches generate rather varied forms of results, \eg 
mere terms in~\cite{gordon1996toward} or oriented multi-graphs 
in~\cite{Cameron2015}. However, we can at least give this basic summary, which 
we corroborate by analysing the actual contents of the results later on. We 
also further discuss the major comparative benefits of our approach in 
Section~\ref{sec:related.lbd}.}. 
In total, we did better than more than half of the related approaches in terms 
of the intermediate ranking. 

The rarity and interestingness measures for the two experiments are given 
in Table~\ref{tab:evaluation_frequency}.
\begin{table}[ht]
\footnotesize
\center
\begin{tabular}{|l||c|c|c|c|}
\hline
Experiment & \multicolumn{2}{|c|}{N} & 
             \multicolumn{2}{|c|}{C~\cite{Cameron2015}} \\
           & $rar(\mathcal{G})$ & $int(\mathcal{G})$ & 
             $rar(\mathcal{G})$ & $int(\mathcal{G})$ \\
\hline\hline
$T_{R}^{40}$ & 6.722 & 0.13  & 0    & 1    \\
\hline
$T_{M}^{39}$ & 0.367 & 0.732 & 0.56 & 0.64 \\
\hline
\end{tabular}
\caption{Claim frequency-based evaluation}\label{tab:evaluation_frequency}
\end{table}
We can only compare ourselves to~\cite{Cameron2015} as the measures 
were defined and used there for the first time. The average results of our
approach are lower than in~\cite{Cameron2015} for the $T_R$ experiment. 
However, the median rarity and interestingness of the paths generated in our 
experiment is 0 and 1, respectively -- only about one third of the $T_R$ path 
associations have non-zero frequency on PubMed. This means that two thirds 
of the claims generated by our approach have the same performance in terms of 
rarity and interestingness as in~\cite{Cameron2015}. The average results of 
the $T_M$ experiment are better in our case. More than 98\% of the $T_M$ 
claims have zero rarity which clearly outperforms~\cite{Cameron2015}. 

Before we continue with the qualitative analysis of the results, let us get 
back for a while to the structural properties of the refined knowledge 
graphs.
Table~\ref{tab:evaluation_degrank} gives average relative ranking of the 
vertices corresponding to the source, target and intermediate terms in the
initially extracted and refined graphs. The rankings are based on the vertex 
degree and betweenness centrality measures (from highest to lowest). These
measures are typically used as an approximation of a vertex importance within
a graph~\cite{sna}.
\begin{table}[ht]
\footnotesize
\center
\begin{tabular}{|l||c|c||c|c|c|c||c|c|}
\hline
Terms          & \multicolumn{4}{|c|}{degree ranking} & 
                 \multicolumn{4}{|c|}{betw. centrality ranking} \\
               &   $T_R^0$   &   $T_R^{40}$  &     $T_M^0$   &  $T_M^{39}$ 
               &   $T_R^0$   &   $T_R^{40}$  &     $T_M^0$   &  $T_M^{39}$ \\
\hline\hline
Source, target & 0.522       & 0.609         & 0.541         & 0.708       
               & 0.537       & 0.592         & 0.656         & 0.736       \\ 
\hline
Intermediates  & 0.563       & 0.729         & 0.557         & 0.57        
               & 0.678       & 0.699         & 0.584         & 0.575       \\ 
\hline
\end{tabular}
\caption{Degree-based ranking of the re-discovery 
terms}\label{tab:evaluation_degrank}
\end{table}
The importance of sources, targets and intermediates in terms of degree is 
increased by the refinement in both experiments. The increase is largest for 
source and target terms in the $T_M$ experiment and for the $T_R$ 
intermediates. The importance in terms of betweenness centrality is increasing
relatively less, with the $T_M$ intermediates actually becoming slightly less
important. These observations are consistent with the evidence-based evaluation
in the sense of ``sensitivity'' of the experimental data sets towards the 
source, target and intermediate terms. The refinement of the $T_R$ graph 
clearly raises the importance of all vertices, especially the intermediates. 
Indeed, all the terms are present in relatively highly ranking claims of the 
resulting $T_R^{40}$ graph. For the $T_M$ data set where only the importance of
the source and target vertices is markedly rising, the results are much 
sparser -- although the $T_M^{39}$ graph contains many claims connecting the 
source and target, there is relatively few intermediates 
from~\cite{swanson1987migraine} present in these claims. 

The qualitative analysis of the solution contents further elaborates on the 
above observations about the initial and resulting graph structure. As 
specified in Section~\ref{sec:evaluation.experiments.evaluation_methodology}, 
the analysis is based on the topics covered by the solution graphs. These are 
terms that provide additional context for the intermediates in the solutions. 
We provide comprehensive lists of the unique context topics in Appendix~A, 
together with references to supporting literature. 

The contents of the Appendix~A is summarised in 
Table~\ref{tab:evaluation_qualitative} which contains the $top_d,\;top_r,\;
top_n$ score values for the initial and refined knowledge graphs in both 
experiments\footnote{As we are not experts in the domains involved, 
we adopted a very conservative strategy for 
determining the topic relevance. If we could not directly verify any 
particular relationship between biomedical concepts present in the solution 
graphs using a review of published literature via PubMed, we asserted the 
corresponding solution irrelevant. We encourage more knowledgeable readers 
to suggest possible updates of the detailed tables in Appendix~A.}. 
\begin{table}[ht]
\footnotesize
\center
\begin{tabular}{|l||c|c||c|c|}
\hline
Score    & \multicolumn{2}{|c||}{$T_R$} & \multicolumn{2}{|c|}{$T_M$} \\
         &   $T_R^0$   &   $T_R^{40}$  &     $T_M^0$   &  $T_M^{39}$ \\
\hline\hline
$top_d$ & 0.412        & 0.522         & 0.368         & 0.818       \\
\hline
$top_r$ & 0.571        & 0.75          & 0.607         & 0.889       \\
\hline
$top_n$ & 0.938        & 0.889         & 0.471         & 0.875       \\
\hline
\end{tabular}
\caption{Summary of the qualitative 
evaluation}\label{tab:evaluation_qualitative}
\end{table}
The table shows that our approach improves the quality of the extracted
knowledge graphs. The relative topical density $top_r$ (\ie the ratio of 
unique topics among the paths connecting source and target terms) increases 
by about 27\% and 122\% for the $T_R, T_M$ experiments, respectively. The
relevance $top_r$ increases by about 31\% and 46\% for $T_R, T_M$, 
respectively. Finally, the relative topical novelty $top_n$ increases by about
86\% for the $T_M$ experiment. In case of the $T_R$ experiment, the measure is 
slightly lower for $T_R^{40}$ than for $T_R^0$, however, there is only one 
non-novel solution in both knowledge graphs. The decrease in the relative 
$top_r$ value is caused by lower total number of solutions in the refined 
graph.

These results 
confirm our 
assumption that the refinement 
improves the quality of statements
extracted from literature, at least in the context of two standard 
literature-based discovery scenarios. The improvement in quality is three-fold. 
Firstly, the refined knowledge graphs are less redundant (the topical
density is higher). Secondly, there is markedly more relevant solutions in the 
results. And thirdly, the refined solutions are largely non-obvious (high 
$top_n$ measure). 

\begin{figure}
\center
\begin{sideways}
  \begin{minipage}{17.5cm}
    \scalebox{0.55}{\includegraphics{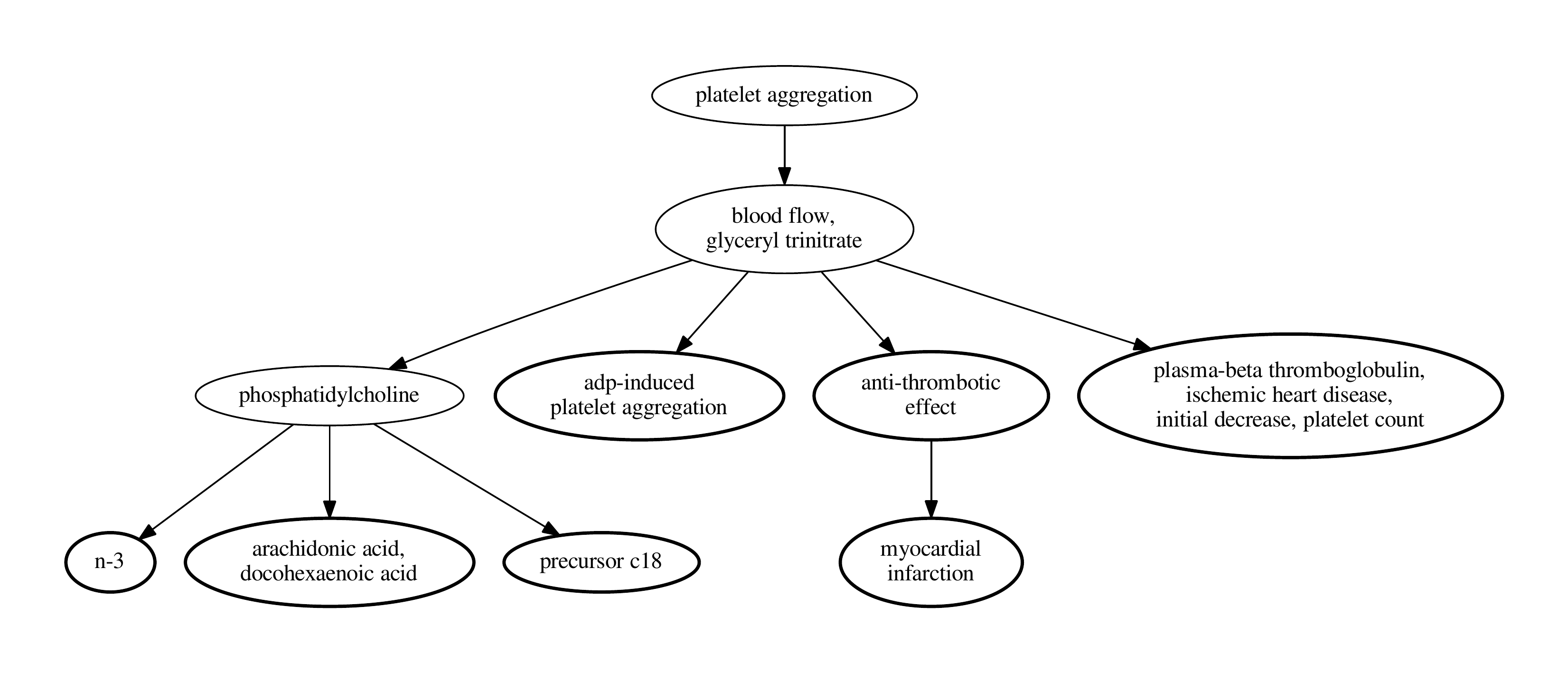}}\\
    \scalebox{0.55}{\includegraphics{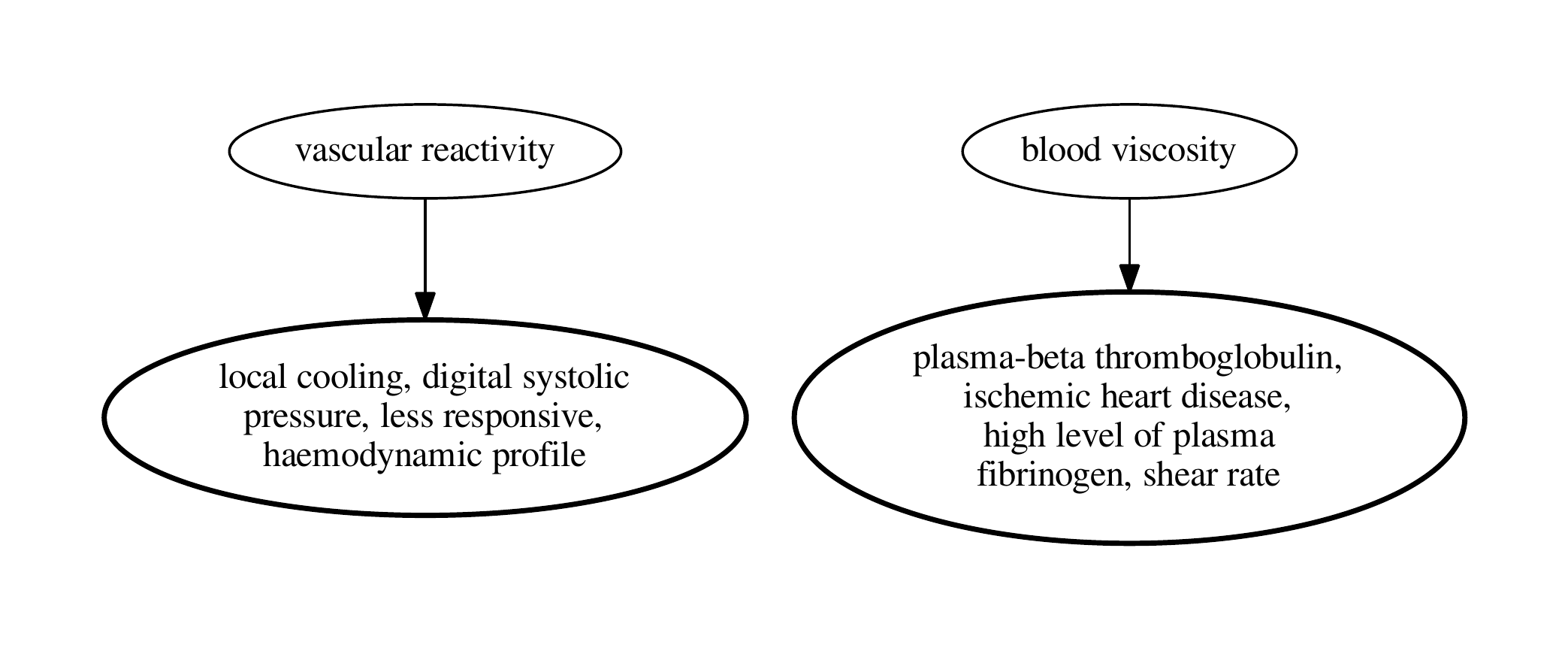}}
  \end{minipage}
\end{sideways}
\caption{Hierarchy of relevant topics in $T_R^{40}$}\label{fig:topics_r}
\end{figure}

Direct and exact comparison of our qualitative results to related state of the 
art is unfortunately impossible due to the afore-mentioned differences in the 
solution representations. However, we can at least discuss the commonalities 
and differences informally. Figure~\ref{fig:topics_r} displays the hierarchy 
of topics covered by the $T_R$ results. 
Each vertex in the hierarchy graph represents a part of the topic. The roots
of the presented hierarchies are the intermediate concepts. The vertices shared
across multiple topics have normal outlines. Vertices that complete
the topics on the way from the root have bold outlines. 

The impact of glyceryl trinitrate on vasodilation and consequently also
on blood flow has been studied in the context of possible treatment of 
Raynaud's syndrome~\cite{kleckner1951effect}. Our method reflects these
findings in constructing a corresponding connection between Raynaud's syndrome
and platelet aggregation which is quite closely related to blood 
flow~\cite{jackson2007growing}. Phosphatidylcholine, also a relatively common 
vertex in the generated claims, refers to a class of phospholipids that is 
closely related to metabolism of fatty acids, including those found 
in fish oils~\cite{abe2014novel}. The vertices connected to 
phosphatidylcholine mediate the relationship between fish oil and platelet 
aggregation in the solutions. The topic with ADP-induced platelet 
aggregation~\cite{puri1999adp} specifies the type of platelet aggregation fish 
oils can influence. The solutions concerned with anti-thrombotic effect put 
this vertex in connection with fish oils, possibly with an intermediate vertex 
referring to myocardial infarction. This corresponds to the anti-thrombotic 
effect of fish oils demonstrated for instance in~\cite{zhu1990modification}. 
Finally, one of our solutions identified a link between platelet aggregation 
and fish oils via their influence on levels of plasma-beta thromboglobulin, a 
marker in ischemic heart disease~\cite{hay1982effect}. 

The solution involving the vascular reactivity intermediate puts it in the
context of influence of fish oils on lower vascular resistance, as 
discussed for instance in~\cite{mozaffarian2007fish}. The effect of local 
cooling on the digital systolic pressure is inherent to Raynaud's syndrome
but its connection to the vascular reactivity discovered by our approach is
more indirect. One branch of the solution involving the blood viscosity 
intermediate revisits the relationship between fish oils and ischemic
heart disease observed in one of the claims containing platelet aggregation. 
The other branch is new, though, and puts the blood viscosity in relation with
high levels of fibrinogen in Raynaud's syndrome 
patients~\cite{tietjen1975blood}. 

When comparing the contents of the $T_R$ solutions with related state of the 
art approaches, we can only refer to~\cite{Cameron2015} 
and~\cite{swanson1986fish} as the other works generate mere lists of possible 
intermediates without further context. 
Many contexts associated with the intermediates as possible
explanations of the connections are missing in the related works. Examples are
blood flow, glyceryl trinitrate, ADP-induced platelet aggregation, 
phosphatidylcholine or plasma-beta thromboglobulin within ischemic heart 
disease. However, most of these connections are rather explanatory and not 
essential in the scope of Raynaud's syndrome despite of being valid. 
In~\cite{Cameron2015}, many of the graphs involve epoprostenol (essentially a 
prostaglandin) as a mediator of the influence of fish oils on platelet 
aggregation. This is consistent with~\cite{swanson1986fish} that establishes 
the connection between fish oil and platelet aggregation as a result of 
increased level of prostaglandins. This context is missing in our results that 
involve the intermediates, however, it is present twice among the top-ten 
solutions (at ranks 4 and 8). Once it appears in relation to the action of the 
drug indomethacin, and then also in relation to luteolytic activity in women 
with Raynaud disease. 
These are potentially interesting findings that 
extend the results produced by comparable state of the art approaches. 

\begin{figure}
\center\begin{sideways}
  \begin{minipage}{17.5cm}
    \scalebox{0.55}{\includegraphics{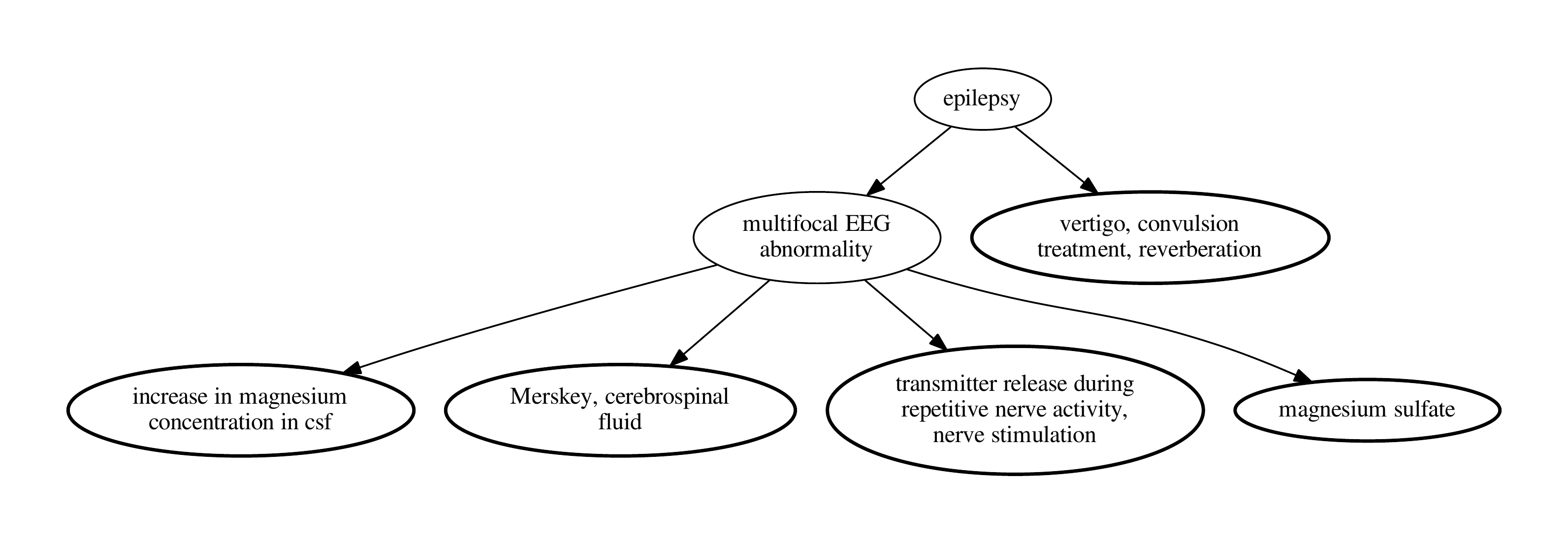}}\\
    \scalebox{0.55}{\includegraphics{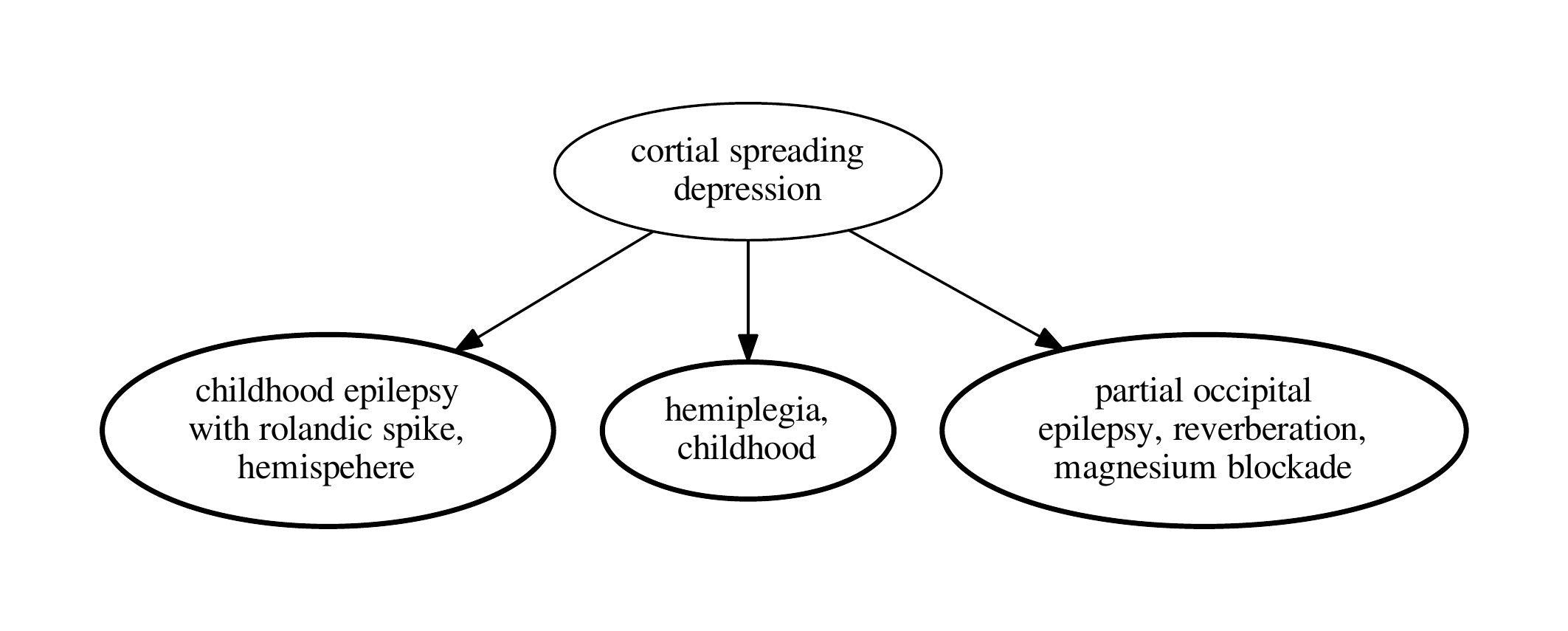}}
  \end{minipage}
\end{sideways}
\caption{Hierarchy of relevant topics in $T_M^{39}$}\label{fig:topics_m}
\end{figure}

Figure~\ref{fig:topics_m} displays the hierarchy of topics covered by the 
$T_M$ results. 
One solution involving the epilepsy intermediate puts it in the context of
magnesium being used as a mechanism for management of reverberating brain 
waves~\cite{shibata1975techniques}. These are associated with epilepsy and
vertigo attacks and the solution suggests that treatments for these conditions
may be used for migraine as well. Other claims related to epilepsy all share
multifocal EEG abnormalities which are characteristic for 
epilepsy~\cite{noriega1976clinical}. Two different types of claims were
complementing these findings -- two solutions dealing with magnesium 
concentrations in cerebrospinal fluid in relation to 
migraine~\cite{ramadan1989low}, and one solution related to transmitter
release and nerve stimulation. 
The solutions involving the cortical spreading depression intermediate were
all related to similar concepts as the epilepsy ones. This is not surprising,
since cortical spreading depression is quite closely related to 
seizures~\cite{fabricius2008association}.

Similarly to the $T_R$ experiment, we can only compare the contents of our 
$T_M$ solutions to~\cite{Cameron2015} and~\cite{swanson1987migraine} which are 
the only works that provide context in addition to the 
intermediates. The graphs presented in~\cite{Cameron2015} for migraine and 
magnesium are generally much sparser than those for Raynaud and fish oil 
(typically containing only the source, target and intermediate node). Moreover,
none of the results discussed in the article in detail concern epilepsy or 
spreading depression. The work~\cite{swanson1987migraine} confirms close 
relationship between epilepsy and cortical spreading depression which is 
consistent with a straightforward interpretation of our results. Our solutions 
also managed to bring up the relationship between magnesium and cerebrospinal 
fluid in the context of epileptic attacks. In addition to that, our results 
appear to strongly associate migraine with multifocal EEG abnormalities. This 
is consistent with relationship between the abnormalities and headaches 
demonstrated for instance in~\cite{guidetti1986headache}. Other 
potentially interesting findings not covered by related works are vertigo, 
reverberation and the relationship between migraine, cortical spreading 
depression and rolandic epilepsy~\cite{wirrell2006children}. 


\section{Related Work}\label{sec:related}

We split this section into four thematic blocks that correspond to the main
theoretical and application-specific facets of our work. In particular, we 
review the areas of:
\begin{inparaenum}[1.]
  \item automated discovery,
  \item ontology learning,
  \item discovery supported by knowledge graphs,
  \item literature-based discovery.
\end{inparaenum}

\subsection{Automated Discovery}\label{sec:related.ad}

Research of ways how discoveries can be automated or facilitated by machines 
dates back to the dawn of the digital computer era. The 
work~\cite{newell1959processes} provides a comprehensive analysis of the 
discovery process operationalised as creative problem solving. It reviews 
several classic machine discovery systems and the heuristics used by them, and
also mentions several properties of worthy discoveries like novelty and value. 
A more recent related work~\cite{klahr1999studies} reviews the major approaches
to studying the process of scientific discovery, provides another survey of 
automated discovery systems and analyses additional features of relevant 
discoveries, such as surprise. The works~\cite{Langley1989283,LANGLEY2000393} 
review still more machine discovery systems and heuristics, 
and identify features like 
refutability and simplicity as essential to discoveries. One of the most recent
and relevant works from this area is~\cite{maher2012using}. It builds 
on~\cite{newell1959processes,klahr1999studies} and introduces 
formalisations of several discovery features. In particular, it models novelty 
and value using metric spaces, and surprise using Bayesian probabilities. 

Discovery features discussed in the referenced works conform to our virtue 
definitions, although most of them do not provide a systematic formalisation, 
only rather application-specific implementations. For instance, refutability 
and simplicity as reviewed in~\cite{Langley1989283} directly correspond to our 
virtues. Surprise and novelty discussed in the other works can be modelled by 
putting emphasis on radical claims as addressed by the conservatism virtue, 
only using different distance metrics for each of the respective features. We 
believe that our approach presents a new way to formalising discovery features 
that is consistent with related state of the art, but is more systematic, 
comprehensive and extensible. In addition, we provide an actionable set of 
measures implemented in the context of knowledge graphs. This enables 
universal applicability of our research, which is not the case in most of the 
rather specific afore-mentioned approaches.

\subsection{Ontology Learning}\label{sec:related.ol}

In the last fifteen years, there has been a growing interest in exploring the 
potential of automatically extracted graph structures for knowledge discovery.
Many of such approaches can be clustered under the umbrella of ontology 
learning~\cite{Maedche2004} which aims at extracting complex statements from 
unstructured textual resources. This is done using specifically tailored 
methods from AI disciplines like natural language processing or machine 
learning. 

As a recent survey~\cite{Wong:2012:OLT:2333112.2333115} shows, the 
applicability of existing ontology learning approaches to (semi)automated 
knowledge discovery is still quite limited. Many of the techniques are 
dependent on manually curated resources. They also introduce a lot of 
assumptions during the extraction process (based on, for instance, linguistic 
facts valid only in the context of a particular language or discourse). This 
limits their universal applicability. Another problem is that the more complex 
knowledge representation the learned ontologies use, the more restrictive they 
are about their meaning. This typically leads to brittleness \wrt 
the often inherently vague and contextual nature of the knowledge they 
represent. This can easily cause problems in machine-aided knowledge discovery 
scenarios where we typically want to represent the knowledge implied by the 
input data in as unbiased way as possible. Another practical limitation is 
that most ontology learning system do not scale very well as reported 
in~\cite{Wong:2012:OLT:2333112.2333115}. 

\subsection{Discovery Supported by Knowledge Graphs}\label{sec:related.kg}

More recent works related to machine discovery using knowledge graphs 
include~\cite{colazzo:hal-00960609,dong2014knowledge,nickel2015review} which 
contain also comprehensive reviews of prior similar approaches. The approach 
elaborated in~\cite{colazzo:hal-00960609} presents methods for knowledge 
discovery in RDF~\cite{RDF} data based on user-defined query patterns and 
analytical perspectives. Our approach complements~\cite{colazzo:hal-00960609} 
by offering means for automated analysis and refinement of knowledge graphs 
using application-independent, well-founded features. 

Google's Knowledge Vault~\cite{dong2014knowledge} presents a web-scale 
approach to probabilistic knowledge fusion that uses graphs represented in the 
RDF format. It tackles the scalability vs. accuracy trade-off of the manual 
and automatic approaches to construction of knowledge graphs. This is done by 
refining statements extracted from the web content using models learned from 
pre-existing highly accurate knowledge bases like YAGO or Freebase. Additional 
details and broader theoretical context of the approach introduced 
in~\cite{dong2014knowledge} is given in~\cite{nickel2015review}, which offers 
a comprehensive review of relational machine learning approaches in the context 
of RDF-compatible knowledge graphs. 
The main advantage of our approach \wrt the 
works~\cite{dong2014knowledge,nickel2015review} is that we are not critically 
dependent on the background knowledge model. In addition, we present a 
complementary well-founded approach to determining which relationships in 
automatically extracted knowledge graphs are worth preservation. Having said 
that, the techniques reviewed in~\cite{nickel2015review} can certainly provide 
valuable hints on future extensions of our approach to graphs with oriented 
edges representing more than one type of relationships (\ie RDF graphs). 

\subsection{Literature-Based Discovery}\label{sec:related.lbd}

As our approach has been validated by experiments in literature-based 
discovery, we need to position ourselves within that field as well. Surveys of 
related works (focusing mostly on the domain of life sciences) are provided 
in~\cite{deBruijn20027,preiss2012towards,smalheiser2012literature}. 
The specific approaches we compare ourselves to are described 
in~\cite{Cameron2015,blake2002automatically,hristovski2006exploiting,
weeber2001using,srinivasan2004text,gordon1996toward}. In most cases where we 
were able to 
directly compare our results with the related works, our approach was at least 
as good as and often better than the state of the art. In addition, we managed 
to hint at several relevant insights that were not even discussed by the human
expert in the original studies~\cite{swanson1986fish,swanson1987migraine}. 

The most significant advantages of our approach are, however, as follows.
\begin{inparaenum}[1.]
  \item It is absolutely automatic. The only manual action we performed was 
  pruning the fulltext search results when mapping terms to the corresponding 
  vertices, however, this is only required for the evaluation, not for the 
  method itself.
  \item There are no domain-specific dependencies and thus our work is readily 
  applicable to any field, not just the biomedical literature-based discovery. 
  \item We produce extensive contextual information that can facilitate the
  interpretation of the results and thus make the machine-aided discovery 
  process more efficient.
  \item Our approach is based on theoretical foundations motivated by the 
  state of the art philosophical study of key features of scientific 
  discoveries.
\end{inparaenum}

The works~\cite{srinivasan2004text,hristovski2006exploiting,weeber2001using}
all depend on rather extensive manual effort (definition of semantic types and
discovery patterns, result pruning, {\it etc.}). The 
approaches~\cite{Cameron2015,blake2002automatically,gordon1996toward} are 
automated, however, 
only~\cite{Cameron2015} provides broader context in order to elucidate the 
connections. Moreover, all works but~\cite{gordon1996toward} substantially 
rely on an external domain-specific source of background knowledge and/or 
domain-specific NLP tools, namely the MeSH and UMLS 
vocabularies~\cite{bodenreider2004unified} and the tools 
SemRep~\cite{rindflesch2005semantic} and BioMedLEE~\cite{chen2004extracting}. 
It is quite plausible to assume that without these resources, the related 
approaches dependent on them would perform much less favourably when compared 
to our implementation. Last but not least, all the related approaches lack 
the universally applicable theoretical foundations presented as the core
contribution of this article.


\section{Conclusions and Future Work}\label{sec:conclusion}

We have presented a novel approach to discovery informatics that is based on
formalisation of hypothesis virtues in the context of knowledge graphs. We have
shown that the approach is naturally motivated, well-founded, extensible and 
universally applicable. It can be used as a broader theoretical frame for other
approaches to machine discovery, as briefly outlined in 
Section~\ref{sec:related}. We have delivered an implementation of the 
presented research and performed its experimental validation using standard 
scenarios in literature-based discovery. A successful comparison with related 
state of the art tools demonstrates the practical relevance of our work.

In near future, we will extend the theoretical framework in order to address 
directed multi-graphs with predicate edge labels and more complex semantics 
associated with particular edge and vertex types. This will allow for 
straightforward application of our approach to more expressive knowledge 
graphs, such as RDF~\cite{RDF} knowledge bases and ontologies in the Linked 
Open Data cloud~\cite{ldbook}. 
Furthermore, we intend to continue demonstrating the universality of our 
framework by using it in other experimental scenarios targeted by related 
works in machine discovery. We also plan to explore the complex relationships 
between specific measures and their influence on the properties of the 
evolutionary refinement process (\eg convergence, optimality and completeness
bounds). This will lead to deeper understanding of the refinement, and 
therefore also to more efficient implementations. Finally, and perhaps most 
importantly, we would like to use our approach in scenarios involving actual 
new discoveries, in direct collaboration with corresponding domain experts.



\section*{References}
\bibliographystyle{elsarticle-harv.bst}
\bibliography{vit.bib}

\section*{Appendix A: Context Topics for the Intermediate Terms}

This appendix presents tables with the context topics of the intermediate
terms for the initial and refined graphs in the $T_R, T_M$ experiments. 
Each table contains topics for one intermediate within a specific 
experimental graph. The topics are given in the first column. The second
column of each table states whether or not the corresponding topics are
relevant. Topics with relevant part subsumed by another relevant topic are 
not considered relevant unless they extend the subsumed part with new relevant 
information. If there is an exclamation mark in the second column (applicable
only to relevant topics), it means the relationship expressed by the 
corresponding solution is novel, \ie not covered in any single existing 
article published by March, 2015. The third column lists references of 
articles that jointly provide support of relevant claims. We use easily 
de-referencable PubMed identifiers for brevity. Note that more common and/or 
simple relationships may have alternative sets of supporting articles that do 
not appear in the lists provided here. 


First we provide three tables for the initial graph $T_R^{0}$ in the $T_R$
experiment 
(Table~\ref{tab:evaluation_topics.tr.0.1}-\ref{tab:evaluation_topics.tr.0.3})
and three tables for the refined graph $T_R^{40}$
(Table~\ref{tab:evaluation_topics.tr.40.1}-\ref{tab:evaluation_topics.tr.40.3}).

\begin{table}
\footnotesize
\center
\begin{tabular}{|p{0.5\linewidth}|c|p{0.35\linewidth}|}
\hline
Topics                                               & Rel. & Support        \\
\hline\hline
glyceril trinitrate, blood flow, myocardial
infarction                                           & Y!   & 17311994, 14831190, 10604966, 3092847, 9310278       \\
\hline
glyceril trinitrate, blood flow, precursor c18, c20  & N    & N/A            \\
\hline
glyceril trinitrate, blood flow, phosphatidylcholine,
arachidonic acid                                     & Y!   & 17311994, 9675609, 4205973, 14831190, 10604966, 3092847        \\
\hline
glyceril trinitrate, blood flow, aa and 
docosahexaenoic acid                                 & N    & N/A            \\
\hline
glyceril trinitrate, blood flow, phosphatidylcholine, 
individual phospholipid, alteration and recovery     & N    & N/A            \\
\hline
glyceril trinitrate, blood flow, antithrombin iii    & Y!   & 17311994, 18370504, 14831190, 10604966, 3092847               \\
\hline
glyceril trinitrate, blood flow, adp-induced platelet 
aggregation                                          & Y!   & 17311994, 10086317, 18370504, 14831190, 10604966, 3092847     \\
\hline
glyceril trinitrate, blood flow, adp-induced platelet 
aggregation, corn oil                                & N    & N/A            \\
\hline
glyceril trinitrate, blood flow, linseed             & N    & N/A            \\
\hline
glyceril trinitrate, blood flow, thrombin and 
collagen                                             & Y!   & 17311994, 4468230, 18370504, 14831190, 10604966, 3092847               \\
\hline
glyceril trinitrate, blood flow, collagen, precursor
c18                                                  & N    & N/A            \\
\hline
glyceril trinitrate, blood flow, adp-induced platelet
aggregation, vasospastic disease, prostaglandin e1   & Y!   & 17311994. 10086317, 18370504, 14831190, 10604966, 3092847               \\
\hline
glyceril trinitrate, blood flow, low pufa coconut oil
adp--induced platelet aggregation                    & N    & N/A            \\
\hline
alpha-adrenergic receptor                            & N    & N/A            \\
\hline
prostacyclin, thromboxane, adp-induced platelet 
aggregation                                          & Y!   & 10086317, 6258879, 19037602     \\
\hline
upper limb, arteritis, haemodynamic profile          & N    & N/A            \\
\hline
alpha-adrenergic receptor, adp-induced platelet 
aggregation                                          & Y    & 34707         \\
\hline
\end{tabular}
\caption{Topic contexts for {\bf platelet aggregation} in 
$T_R^{0}$}\label{tab:evaluation_topics.tr.0.1}
\end{table}

\begin{table}
\footnotesize
\center
\begin{tabular}{|p{0.5\linewidth}|c|p{0.35\linewidth}|}
\hline
Topics                                               & Rel. & Support        \\
\hline\hline
glyceryl trinitrate, blood flow, hand, cold          & N    & N/A            \\
\hline
glyceryl trinitrate, blood flow, hand, cold, 
myocardial infarction                                & Y!   & 24753696, 9310278, 10086317, 18370504, 14831190        \\
\hline
vascular dilation, cold, myocardial infarction       & Y!   & 9310278, 15695304      \\
\hline
upper limb, arteritis, haemodynamic profile, 
myocardial infarction                                & N    & N/A            \\
\hline
local cooling, digital systolic pressure, 
haemodynamic profile, less responsive,
myocardial infarction                                & Y!   & 9310278, 22453196, 17876193       \\
\hline
cold spell, radiological investigation               & N    & N/A            \\
\hline
\end{tabular}
\caption{Topic contexts for {\bf vascular reactivity} in 
$T_R^{0}$}\label{tab:evaluation_topics.tr.0.2}
\end{table}

\begin{table}
\footnotesize
\center
\begin{tabular}{|p{0.5\linewidth}|c|p{0.35\linewidth}|}
\hline
Topics                                               & Rel. & Support        \\
\hline\hline
finger systolic pressure, dazoxiben treatment, high
level of plasma fibrinogen                           & Y!   & 6393521, 7459607 \\
\hline
fibrinolytic enhancement, high level of plasma 
fibrinogen                                           & Y!   & 698554, 7459607 \\
\hline
fibrinolytic enhancement, deformability index        & Y!   & 698554, 519354  \\
\hline
finger systolic pressure, dazoxiben treatment, high
level of plasma fibrinogen, shear rate               & Y!   & 6393521, 7459607, 579511               \\
\hline
finger systolic pressure, dazoxiben treatment, high
level of plasma fibrinogen, shear rate, fibrinolytic 
enhancement, deformability index                     & Y!   & 6393521, 7459607, 579511, 519354               \\
\hline
\end{tabular}
\caption{Topic contexts for {\bf blood viscosity} in 
$T_R^{0}$}\label{tab:evaluation_topics.tr.0.3}
\end{table}

\begin{table}
\footnotesize
\center
\begin{tabular}{|p{0.5\linewidth}|c|p{0.35\linewidth}|}
\hline
Topics                                               & Rel. & Support        \\
\hline\hline
n-3, phosphatidylcholine, blood flow, glyceryl 
trinitrate                                           & Y!   & 17311994, 9675609, 24679762, 14831190, 10604966, 3092847               \\
\hline
phosphatidylcholine, aa and docohexaenoic acid,
arachidonic acid, blood flow, glyceryl trinitrate    & Y!   & 17311994, 9675609, 4205973, 14831190, 10604966, 3092847               \\
\hline
plasma-beta-thromboglobulin, ischemic heart disease,
platelet count, initial decrease, blood flow, 
glyceryl trinitrate                                  & Y!   & 17311994, 6123019, 14831190, 10604966, 3092847               \\ 
\hline
adp-induced platelet aggregation, blood flow, 
glyceryl trinitrate                                  & Y!   & 17311994, 10086317, 18370504, 14831190, 10604966, 3092847               \\
\hline
anti-thrombotic effect, blood flow, glyceryl 
trinitrate                                           & Y!   & 17311994, 6294902, 14831190, 10604966, 3092847               \\
\hline
anti-thrombotic effect, myocardial infarction, blood 
flow, glyceryl trinitrate                            & Y!   & 17311994, 6294902, 9310278, 14831190, 10604966, 3092847               \\
\hline
n-3, phosphatidylcholine, blood flow, glyceryl 
trinitrate, alteration and recovery                  & N    & N/A            \\
\hline
n-3, phosphatidylcholine, blood flow, glyceryl 
trinitrate, alteration and recovery, individual
phospholipid                                         & N    & N/A            \\
\hline
precursor c18, phosphatidylcholine, blood flow,
glyceryl trinitrate                                  & Y!   & 17311994, 9675609, 4205973, 14831190, 10604966, 3092847 \\
\hline
\end{tabular}
\caption{Topic contexts for {\bf platelet aggregation} in 
$T_R^{40}$}\label{tab:evaluation_topics.tr.40.1}
\end{table}

\begin{table}
\footnotesize
\center
\begin{tabular}{|p{0.5\linewidth}|c|p{0.35\linewidth}|}
\hline
Topics                                               & Rel. & Support        \\
\hline\hline
local cooling, digital systolic pressure, 
less responsive, haemodynamic profile                & Y    & 22453196, 17876193       \\
\hline
upper limb, arteritis, haemodynamic profile,
myocardial infarction                                & N    & N/A            \\
\hline
\end{tabular}
\caption{Topic contexts for {\bf vascular reactivity} in 
$T_R^{40}$}\label{tab:evaluation_topics.tr.40.2}
\end{table}

\begin{table}
\footnotesize
\center
\begin{tabular}{|p{0.5\linewidth}|c|p{0.35\linewidth}|}
\hline
Topics                                               & Rel. & Support        \\
\hline\hline
plasma-beta-thromboglobulin, ischemic heart disease, 
high level of plasma fibrinogen, shear rate          & Y!   & 6123019, 579511, 519354         \\ 
\hline
\end{tabular}
\caption{Topic contexts for {\bf blood viscosity} in 
$T_R^{40}$}\label{tab:evaluation_topics.tr.40.3}
\end{table}


The topics for the $T_M$ experiment are organised in five tables for the 
initial graph $T_M^{0}$ 
(Table~\ref{tab:evaluation_topics.tm.0.1}-\ref{tab:evaluation_topics.tm.0.5})
and in three tables for the refined graph $T_M^{39}$
(Table~\ref{tab:evaluation_topics.tm.39.1}-\ref{tab:evaluation_topics.tm.39.3}).

\begin{table}
\footnotesize
\center
\begin{tabular}{|p{0.5\linewidth}|c|p{0.35\linewidth}|}
\hline
Topics                                               & Rel. & Support        \\
\hline\hline
acute intrapartum fetal distress                     & N    & N/A            \\
\hline
flunarizine, seizure                                 & Y    & 25754865, 3332609, 22406257               \\
\hline
toxemia, horton                                      & Y!   & 25754865, 25373431, 14401618, 6557975       \\
\hline
bulbo-cortical pathway                               & Y    & 25754865, 3609865        \\
\hline
sea water                                            & N    & N/A            \\
\hline
m.e.p.p., inhibitory effect of prostaglandin on
vasopressin                                          & N    & N/A            \\
\hline
brachymetapody, inhibitory effect of prostaglandin 
on vasopressin, cyanosis                             & N    & N/A           \\
\hline
fibrinolysis, lipid, vascular disease                & Y!   & 25754865, 20669129, 25737193     \\
\hline
antiserotonin, toxemia                               & Y!   & 25754865, 25373431, 5315778                \\
\hline
spasmophilia, magnesium sulfate                      & Y    & 25754865, 9340190        \\
\hline
cyanosis, limb, inhibitory effect of prostaglandin
on vasopresin, magnesium excretion                   & N    & N/A            \\
\hline
conventional therapy for vertigo                     & Y    & 25754865, 23837033               \\
\hline
amonium                                              & Y    & 25754865, 10897167               \\
\hline
verapamil, magnesium blockade                        & Y!   & 25754865, 23973639, 24113539, 8891316               \\
\hline
increase in bathing, washing                         & Y!   & 25754865, 2294020025667882               \\
\hline
ketonic body                                         & Y    & 25754865, 24300035               \\
\hline
merskey, cerebrospinal fluid                         & Y    & 25754865, 6100318               \\
\hline
\end{tabular}
\caption{Topic contexts for {\bf epilepsy} in 
$T_M^{0}$}\label{tab:evaluation_topics.tm.0.1}
\end{table}

\begin{table}
\footnotesize
\center
\begin{tabular}{|p{0.5\linewidth}|c|p{0.35\linewidth}|}
\hline
Topics                                               & Rel. & Support        \\
\hline\hline
alkaline phosphatase, hydroxyproline                 & N    & N/A            \\
\hline
magnesium chloride                                   & Y!   & 25238714, 25010639, 24828386,        \\
\hline
immunoreactive, hyposmotic stress                    & N    & N/A            \\
\hline
\end{tabular}
\caption{Topic contexts for {\bf prostaglandin} in 
$T_M^{0}$}\label{tab:evaluation_topics.tm.0.2}
\end{table}

\begin{table}
\footnotesize
\center
\begin{tabular}{|p{0.5\linewidth}|c|p{0.35\linewidth}|}
\hline
Topics                                               & Rel. & Support        \\
\hline\hline
magnesium blockade, benign syndrome                  & N    & N/A            \\
\hline
\end{tabular}
\caption{Topic contexts for {\bf cortical spreading depression} in 
$T_M^{0}$}\label{tab:evaluation_topics.tm.0.3}
\end{table}

\begin{table}
\footnotesize
\center
\begin{tabular}{|p{0.5\linewidth}|c|p{0.35\linewidth}|}
\hline
Topics                                               & Rel. & Support        \\
\hline\hline
indomethacin                                         & Y    & 22529203, 2925371        \\
\hline
high calcium                                         & Y!   & 12010379, 2925371, 15152357               \\
\hline
\end{tabular}
\caption{Topic contexts for {\bf vascular reactivity} in 
$T_M^{0}$}\label{tab:evaluation_topics.tm.0.4}
\end{table}

\begin{table}
\footnotesize
\center
\begin{tabular}{|p{0.5\linewidth}|c|p{0.35\linewidth}|}
\hline
Topics                                               & Rel. & Support        \\
\hline\hline
fatty acid level, aspirin, dipyridamole, epilepsy    & Y!   & 25741817, 25116182, 10775263, 22749692               \\
\hline
cholesterol                                          & N    & N/A            \\
\hline
endocarditis, vasculomotor reaction, epilepsy        & Y!   & 3513926, 21762000               \\
\hline
prostaglandin synthesis, potent anti-inflammatory
agent                                                & N    & N/A            \\
\hline
\end{tabular}
\caption{Topic contexts for {\bf platelet aggregation} in 
$T_M^{0}$}\label{tab:evaluation_topics.tm.0.5}
\end{table}

\begin{table}
\footnotesize
\center
\begin{tabular}{|p{0.5\linewidth}|c|p{0.35\linewidth}|}
\hline
Topics                                               & Rel. & Support        \\
\hline\hline
transmitter release during repetitive nerve 
activity, nerve stimulation, multifocal EEG 
abnormality                                          & Y!   & 25754865, 2446959, 13175198               \\
\hline
multifocal EEG abnormality, magnesium sulfate        & Y!   & 25754865, 2446959, 23256267               \\
\hline
vertigo, convulsion treatment, reverberation         & Y!   & 25754865, 23837033               \\
\hline
multifocal EEG abnormality, merskey, cerebrospinal
fluid                                                & Y!   & 25754865, 2446959, 6100318               \\
\hline
multifocal EEG abnormality, plasma, increase in 
magnesium concentration in csf                       & Y!   & 25754865, 2446959, 3981211               \\
\hline
\end{tabular}
\caption{Topic contexts for {\bf epilepsy} in 
$T_M^{39}$}\label{tab:evaluation_topics.tm.39.1}
\end{table}

\begin{table}
\footnotesize
\center
\begin{tabular}{|p{0.5\linewidth}|c|p{0.35\linewidth}|}
\hline
Topics                                               & Rel. & Support        \\
\hline\hline
childhood epilepsy with rolandic spike, hemisphere   & Y!   & 25754865, 19271946, 19674062, 22961355               \\
\hline
hemiplegia, childhood                                & Y    & 25754865, 23907418, 21490217               \\
\hline
partial occipital epilepsy, reverberation, magnesium
blockade                                             & Y!   & 25754865, 23907418, 1283483, 1110377               \\
\hline
\end{tabular}
\caption{Topic contexts for {\bf cortical spreading depression} in 
$T_M^{39}$}\label{tab:evaluation_topics.tm.39.2}
\end{table}

\begin{table}
\footnotesize
\center
\begin{tabular}{|p{0.5\linewidth}|c|p{0.35\linewidth}|}
\hline
Topics                                               & Rel. & Support        \\
\hline\hline
high calcium, tension headache                       & N    & N/A            \\
\hline
\end{tabular}
\caption{Topic contexts for {\bf vascular reactivity} in 
$T_M^{39}$}\label{tab:evaluation_topics.tm.39.3}
\end{table}

\end{document}